\documentclass[pdflatex,sn-mathphys-num]{sn-jnl}

\usepackage{graphicx}%
\usepackage{multirow}%
\usepackage{amsmath,amssymb,amsfonts}%
\usepackage{amsthm}%
\usepackage{mathrsfs}%
\usepackage[title]{appendix}%
\usepackage{xcolor}%
\usepackage{textcomp}%
\usepackage{manyfoot}%
\usepackage{booktabs}%
\usepackage{algorithm}%
\usepackage{algorithmicx}%
\usepackage{algpseudocode}%
\usepackage{listings}%
\usepackage{comment}%
\usepackage{lmodern}

\theoremstyle{thmstyleone}%
\newtheorem{theorem}{Theorem}
\newtheorem{proposition}[theorem]{Proposition}%

\theoremstyle{definition}

\theoremstyle{thmstyletwo}%

\theoremstyle{thmstylethree}%
\usepackage{microtype}
\usepackage{graphicx}
\usepackage{dsfont}
\usepackage{stmaryrd}
\usepackage{booktabs} 
\usepackage{tikz}
\usetikzlibrary{patterns}
\usetikzlibrary{arrows,calc,shapes,decorations.pathmorphing}

\usepackage{placeins} 
\usetikzlibrary{fit}
\usepackage{boldline}

\usepackage{amsmath}
\usepackage{amssymb}
\usepackage{mathtools}
\usepackage{amsthm}
\usepackage[inkscapelatex=false]{svg}
\usepackage{ulem}
\usepackage{mathrsfs}

\usepackage[capitalize,noabbrev]{cleveref}

\usepackage[all]{hypcap} 
\makeatletter

\usepackage{subcaption}
\makeatother

\newcommand{\RR} {\mathbb R}

\newcommand{\NN} {\mathbb N}

\newcommand{\calQ} {\mathcal Q}
\newcommand{\QQk}[1] {\text{Q}_{#1}}
\newcommand{\QQ} {\text{Q}}

\newcommand {\lb} { \llbracket}      %
\newcommand {\rb} { \rrbracket}      %
\newcommand {\mytop} {\prime} 
\newcommand {\STIEFEL}[1] {St(#1)}      %
\newcommand {\PROJUNN} {P_{\text{projUNN}}}      %
\newcommand {\BJORCK} {P_{\text{Björck}}}      %
\newcommand {\Max} {\alpha}      %

\DeclareMathOperator{\diag}{diag}

\DeclareMathOperator{\FP}{FP}
\DeclareMathOperator{\fp}{fpp}

\newcommand{\myparagraph}[1]{{\bf #1}}

\newcommand{\francois}[1]{\textcolor{blue}{#1}}
\newcommand{\blue}[1]{\textcolor{blue}{#1}}

\begin{document}

\title[On Orthogonal Recurrent Neural Networks quantization]{On Orthogonal Recurrent Neural Networks Quantization}

\author[1]{\sur{Armand} \fnm{Foucault}} \email{firstname.lastname@math.univ-toulouse.fr}

\author[2]{\sur{Franck} \fnm{Mamalet}}
\email{\\ or \qquad \qquad firstname.lastname@irt-saintexupery.com}
\author[1]{\sur{François} \fnm{Malgouyres}}

\affil[1]{Institut de Mathématiques de Toulouse, UMR5219. Université de Toulouse, CNRS. UPS IMT,
F-31062 Toulouse Cedex 9, France}
\affil[2]{Institut de Recherche Technologique Saint Exupéry, Toulouse, France}

\hypersetup{
pdftitle={On Orthogonal Recurrent Neural Networks quantization},
pdfsubject={},
pdfauthor={Armand Foucault, Franck Mamalet, François Malgouyres},
pdfkeywords={Recurrent Neural Networks, Orthogonal Recurrent Neural Networks, Quantized Neural Networks, time series},
}


\abstract{In recent years, Orthogonal Recurrent Neural Networks (ORNNs) have gained popularity due to their ability to manage tasks involving long-term dependencies, such as the Copy-task, and their linear complexity. However, existing ORNNs utilize full precision weights and activations, which prevents their deployment on compact devices.

In this paper, we explore the quantization of the weight matrices in ORNNs, leading to Quantized approximately Orthogonal RNNs (QORNNs). The construction of such networks remained an open problem, acknowledged for its inherent instability. We propose and investigate two strategies to learn QORNN  by combining quantization-aware training (QAT) and orthogonal projections. We also study post-training quantization of the activations for pure integer computation of the recurrent loop. The most efficient fully-quantized models are currently the only alternatives for solving the Copy-task with 1000 timesteps. They also surpass existing quantized models on tasks requiring long-term dependencies, such as permuted MNIST as well as some Internet of Things (IoT) tasks. However, they are less effective on tasks where long short-term memory (LSTMs) and other gated RNNs outperform ORNNs.

}




\maketitle

\section{Introduction}

\label{introduction}
\myparagraph{Motivation:}
Machine learning applications frequently encompass the analysis of time series data, such as textual information and audio signals. Within the realm of deep learning, various Recurrent Neural Network (RNN) architectures and transformers have demonstrated notable success in addressing a diverse array of tasks associated with time series data. 

These models typically require a substantial number of parameters for optimal performance and involve numerous matrix-vector multiplications during inference, using matrices and vectors of considerable sizes containing floating-point numbers. This does not allow for the deployment of these networks on compact devices with memory and power constraints, as well as for real-time applications. Overcoming these constraints allows the use of RNNs across a large range of domains in edge-Machine Learning and tiny-Machine learning applications, as described in \cite{abadade2023comprehensive}, such as healthcare, smart farming, environment, or anomaly
detection.

An effective and often unavoidable step\footnote{For instance, the weights in a Google EdgeTPU are typically encoded as fixed-point numbers using $8$ bits.} to address these challenges is neural network quantization. This technique aims to reduce the number of bits required to represent the weights and activations of the network. As evaluated in \cite{hubara2018quantized,gholami2022survey}, with appropriate hardware and implementation, this accelerates runtime computations, lowers power consumption, and, on the other hand, decreases the amount of space needed for parameter storage. Other authors have gone further and implemented quantized LSTMs on low-cost Field-Programmable Gate Arrays (FPGAs) to meet low-power requirements and provide real-time low-cost solutions, see \cite{chen2024eciton,bartels2023integer}.\footnote{In these two references, the only compression of the weights is due to quantization.}


Our goal is to contribute to the field of quantization of RNNs, with a particular emphasis on considering RNNs able to address tasks involving long-term dependencies such as the Copy-task with many time steps. In doing so, we broaden the scope of applications for quantized RNNs.

To achieve this objective, we introduce and compare two strategies for constructing Orthogonal or approximately Orthogonal Neural Networks (ORNNs) with quantized weights. We call them Quantized Orthogonal Recurrent Neural Networks (QORNNs). 
The orthogonality constraint of ORNNs, which has been studied in the articles described in the \Cref{biblio-fp-sec,biblio-URNN-sec}, has indeed the following advantages:
\begin{itemize}
\item Their memory complexity does not increase with the input number of time steps, and their computational complexity scales linearly with the input number of time steps. They are smaller compared to their competitors, which is ideal for training and inference on long inputs.
\item They are easy to learn and have excellent memorization ability, which enables them to efficiently solve important tasks with long-term dependencies, such as the Copy-task.
\end{itemize}
 On the contrary, LSTM and Gated Recurrent Unit (GRU) are known to struggle to solve the Copy-task with many time steps: performance comparable to a naive baseline, consisting of random guessing, have been reported in \cite{Arjovsky2015UnitaryER,bai2018empirical,tallec2018can,kerg2019non,bai2019deep}. For the Copy-task studied in \cite{jelassi2024repeat}, the limitation comes rapidly as the number of time steps of the sequence increases.

\myparagraph{Contribution:}
This paper presents pioneering work in the exploration of the quantization of ORNNs. Our main contributions can be summarized as follows:
\begin{itemize}
    \item We investigate the factors influencing the impact of quantization on orthogonality and the behavior of ORNNs.
   \item We propose two different Quantization-Aware Training (QAT) strategies for constructing ORNNs with quantized weights, called  Quantized and approximately Orthogonal Recurrent Neural Networks (QORNN).
   \item We demonstrate that QORNNs are  quantized recurrent solution that can effectively capture long-term dependencies in the Copy-task with $T>1000$ using only 5 bits for the weights.
   \item We achieve state-of-the-art results on the permuted pixel-by-pixel MNIST (pMNIST) task, even with $4$-bit quantization. 
   \item On GLUE benchmarks, QORNN achieves performance comparable to the smallest quantized transformer, despite being substantially smaller in size.
   \item On IoT benchmarks, QORNN can outperform the state of the art in certain cases.
  \item We further expand our investigation by applying a simple Post-Training Quantization (PTQ) method to the activations, reducing them to 12 bits without any loss in performance. Consequently, we introduce the first fully quantized recurrence capable of solving the Copy-task with sequences longer than $1000$ steps. 
\end{itemize}



\myparagraph{Organization of the paper:} We discuss the related works in \cref{biblio-QRNN}. The notations and technical descriptions related to vanilla RNNs, orthogonality, and quantization are presented in Sections \ref{vRRNs-sec}, \ref{ortho-sec}, and \ref{quant-sec}, respectively. The reasons why quantizing ORNN is unstable are described in \cref{QORNN_hard-sec}. This section also contains bounds on the orthogonality discrepancy of the quantization of orthogonal matrices. The two proposed algorithms for building QORNNs are detailed in \cref{model-sec}. 
We describe our strategy to quantize the activations of the network and obtain a fully quantized RNN in \Cref{add-fullquant}.
Finally, experiments and their results are presented in \cref{exp-sec}. Additional details can be found in the appendices. The code implementing the experiments is available in the \href{https://github.com/deel-ai-papers/QORNN}{GitHub QORNN repository}.

\section{Related works}\label{biblio-QRNN}
In this section, we discuss  full-precision models for time-series, having a linear complexity, in  \Cref{biblio-fp-sec}. We also offer a comprehensive overview of contributions related to Unitary and orthogonal RNNs \footnote{Given that ORNNs achieve comparable performance to URNN \citep{DBLP:conf/icml/MhammediHRB17}, in the scope of lower complexity, we limit this study to ORNN.} in \Cref{biblio-URNN-sec}. We describe works dedicated to quantization methods designed for networks manipulating time-series data in \Cref{related_workquantized_RNN} and provide their results on a selection of benchmarks in \Cref{results_biblio-tab}. 
We also provide bibliographical pointers related to quantized transformers 
and quantized SSMs in \Cref{biblio-quantized-SSMs-sec}.

\subsection{Neural Networks for time series}\label{biblio-fp-sec}

Numerous neural network architectures have been developed specifically for handling time-series data. Rather than attempting to provide an exhaustive overview of all these architectures, our focus is on contextualizing Orthogonal and Unitary Recurrent Neural Networks (ORNN) within this diverse landscape. A comprehensive bibliography on ORNN is deferred to Section \ref{biblio-URNN-sec}.

{\it Vanilla RNNs} as introduced by \cite{werbos1988generalization} are notoriously challenging to train due to issues with vanishing and exploding gradients,  \cite{werbos1988generalization,279181}. This reduces their ability to cope with long-term dependencies. As described in \cref{biblio-QRNN} and reported in \cite{Ott2016RecurrentNN,hubara2018quantized,gupta2022compression}, quantizing Vanilla RNN is challenging and not efficient.

{\it LSTM} \cite{hochreiter1997long} and {\it GRU} \cite{cho-etal-2014-properties} are recurrent architectures that tackle vanishing gradient problems by incorporating gating mechanisms.
They have demonstrated outstanding performance in tasks such as speech recognition \cite{Graves2013SpeechRW} and neural machine translation \cite{Sutskever2014SequenceTS}. LSTM and GRU are known to struggle to solve certain tasks with long-term dependencies, such as the Copy-task with many time steps. On the latter task, comparable performance to a naive baseline, consisting of random guessing, has been reported in \cite{Arjovsky2015UnitaryER,bai2018empirical,tallec2018can,kerg2019non,bai2019deep}. For the Copy-task studied in \cite{jelassi2024repeat}, the limitation comes rapidly as the number of time steps of the sequence increases. To the best of our knowledge, these are the only studies tackling the Copy-task using LSTM and GRU architectures. Several variants of quantized LSTM and GRU have been studied in the literature, see \cref{related_workquantized_RNN}.

{\it ORNN (and URNN)} are known to achieve superior performance compared to LSTM in handling time series with long-term dependencies \cite{Arjovsky2015UnitaryER}. They explicitly address the issues of vanishing and exploding gradients by imposing orthogonality (and unitary) constraints on the recurrent weight matrix of a vanilla RNN, See \cref{biblio-URNN-sec}. As a bonus, a standard ORNN unit contains about four times fewer parameters than an LSTM unit. To the best of our knowledge, quantization of ORNN has not been previously investigated. A detailed bibliography on full-precision ORNN is in \cref{biblio-URNN-sec}.

Alternative architectures introduce a {\it skip connection} in the RNN, similar to the one in ResNet architectures \cite{he2016deep}. Several studies have contributed to the development of this idea \cite{jaeger2007optimization,bengio2013advances,chang2017dilated}. To the best of our knowledge, the only compressed --and therefore quantized-- version of this architecture is described in the study by \cite{kusupati2018fastgrnn}.

Due to their ability to capture long-term dependencies and their flexibility, {\it Transformer architectures} \cite{Vaswani2017AttentionIA} have demonstrated great performance even for long sequences. However, they often require large training datasets and entail computational complexities that render them unsuitable for this study. Several approaches to Transformer quantization have been proposed in the literature; these works are discussed in \cref{biblio-quantized-transformers-sec}.

Many architectures based on an {\it Ordinary Differential Equation} also benefit from a skip connection and have been studied in several works \cite{chang2018antisymmetricrnn,rusch2020coupled,rusch2021unicornn,lechner2020learning,Kag2020RNNs,kag2021time,erichson2021lipschitz}. Among these, structured state space sequence models (SSM) \cite{gu2021efficiently} have shown effectiveness in handling tasks with very-long-term dependencies, as described in \cite{tay2020long}. 
The quantization of SSMs is also discussed in \cref{biblio-quantized-SSMs-sec}.

Finally, alternatives that do not fit into the categories described above include the Independent RNN \cite{li2018independently}, approaches that utilize alternatives to backpropagation \cite{manchev2020target,ororbia2020continual}, or methods that model infinite-depth networks \cite{bai2019deep}.

To conclude, ORNNs is a method of choice when the two following properties are simultaneously needed:
\begin{itemize}
\item The architecture exhibits memory complexity that is independent of the number of time steps, alongside computational complexity that scales linearly with it. This combination of properties makes it particularly well-suited for inference on tasks involving long input sequences.
\item The architecture is easy to learn and has excellent memorization ability, which permits solving efficiently important tasks with long-term dependencies such as the Copy-task with many time steps.
\end{itemize}

\subsection{Orthogonal and unitary recurrent neural networks}\label{biblio-URNN-sec}

Learning the unitary and orthogonal recurrent weight matrix of recurrent neural networks has a rich history of study in the last decade. We describe the contributions in chronological order of publication. 

Unitary recurrent neural networks (URNNs) were introduced in \cite{Arjovsky2015UnitaryER}. In this article, the recurrent weight matrix is the product of parameterized  unitary matrices of predefined structures. They argue empirically that URNNs are beneficial because they better capture long-term dependencies than LSTMs. Soon after, the authors of \cite{wisdom2016fullcapacity} use the {\it Cayley transform locally} to build an iterative scheme capable of reaching all unitary matrices. In \cite{Jing2016TunableEU}, the authors parameterize the recurrent weight matrix as a product of {\it Givens rotations} and a diagonal matrix. By doing so, they achieve more efficient models with tunable complexity that can be trained more rapidly. In \cite{DBLP:conf/icml/MhammediHRB17}, the authors parameterize the recurrent weight matrix as a product of {\it Householder reflections} to reduce the complexity of full-capacity models. In the same line of research, the authors of \cite{jose2018kronecker} explore the use of {\it product of unitary Kronecker matrices}. They incorporate a soft-orthogonality penalization term to enforce the unitary constraints. The Kronecker architecture can be adjusted to reduce the complexity of the model. In \cite{Vorontsov2017OnOA}, the authors compare soft and hard-orthogonality constraints. They find that the parameter of the soft-orthogonality strategy under study can be tuned to achieve an approximately orthogonal recurrent matrix, leading to improved efficiency. In \cite{helfrich2018orthogonal}, the authors narrow their focus to orthogonal recurrent weight matrices and parameterize the entire Stiefel manifold using the {\it Cayley transform scaled by a diagonal and orthogonal matrix}. Similar to \cite{DBLP:conf/icml/MhammediHRB17}, the number of parameters defining the orthogonal matrix is optimal. The authors of \cite{zhang2018stabilizing} use a parameterized Singular Value Decomposition (SVD) to constrain the singular values of the recurrent weight matrix. In \cite{lezcano2019cheap}, the authors parameterize orthogonal matrices using the {\it exponential map}. Finally, in \cite{kiani2022projunn}, the authors develop two Riemannian optimization strategies. The first one is based on the {\it orthogonal projection onto the Unitary or Stiefel manifold}, and the other on {\it Riemannian geodesic shooting}. The algorithms are named ProjUNN, and one of them is employed in the presented work. This choice is motivated by the experiments outlined in \cite{kiani2022projunn}.

\subsection{On quantized RNNs}\label{related_workquantized_RNN}
The pioneering article on the quantization of weights in RNNs is \cite{Ott2016RecurrentNN}. In this article, the authors explore the quantization of vanilla RNNs, LSTMs, and GRUs. Then, the existing articles consider the quantization of both weights and activations for LSTM \cite{hou2017loss,nia2023training}, LSTM and vanilla RNN \citep{hubara2018quantized} or both LSTM and GRU \citep{zhou2017balanced,ardakani2018learning,xualternating,article-rbn, wang2018hitnet}. The proposals differ in various aspects including the quantization scheme and the optimization strategy. The performances on the most commonly used tasks are summarized in \cref{results_biblio-tab}.
 
 The article \cite{kusupati2018fastgrnn} contains the study of a compressed network named fastRNN whose weights are quantized on $8$ bits and activations on $16$ bits. The architecture of fastRNN contains a skip-connection, similar to the one of ResNet \citep{he2016deep}, and (optionally) a gating mechanism leading to a model called fastGRNN.

 The closest studies evaluate quantized vanilla RNNs; see \cite{Ott2016RecurrentNN} and \cite{hubara2018quantized}; or a subclass of ORNN \cite{foucault2025hadamrnn}. The articles emphasize the difficulty of the problem and only provide results for tasks involving short-term dependencies, such as the next character prediction task on the Penn TreeBank (PTB) and text8 datasets. They explain that this difficulty stems from instability.\footnote{We illustrate and evaluate this phenomenon in \Cref{QORNN_hard-sec}.} The problem of vanilla RNN quantization is also evoked in the recent survey \cite{gupta2022compression}. In \cite{foucault2025hadamrnn}, the authors successfully construct quantized ORNNs. However, compared to the methods described in this paper, their quantization uses only $1$ or $2$ bits. It lacks flexibility, and we will see that it is less efficient on some benchmarks.

Finally, this research contributed to the implementation of quantized RNNs and LSTMs on FPGA, as reported in \cite{chen2024eciton,gao2022spartus,bartels2023integer} and the references therein, leading to a drastic reduction in power consumption, latency, and cost.



\begin{table}[h!]
\centering
\caption{For every article, the best performance is in bold. The Table illustrates that full-precision LSTM is an optimistic surrogate for all quantized RNNs except fastGRNN and fastRNN. We call skip-RNN the network called FastGRNN-LSQ in \cite{kusupati2018fastgrnn}. Weights and activation columns: FP stands for \lq full precision\rq, t for \lq ternary\rq, any number should be interpreted as a bitwidth.}\label{results_biblio-tab}
\begin{tabular}{cccccccc}
\hline
\textbf{Task} &  \textbf{Reference} &\textbf{T}& \textbf{Model} & \textbf{Weights} &\textbf{Activ.}  & \textbf{Score}&\textbf{Metric}\\ \hline

\toprule
PTB & \cite{zhou2017balanced} &---&  LSTM & FP & FP  &  {\bf 97} & PPW\\ 
\cmidrule{5-7}
word  & &  &  & $2$ & $3$ & $123$  &{\footnotesize(lower is better)} \\
\cmidrule{4-7}
 & &&  GRU & FP & FP  &  $100$&  \\
\cmidrule{5-7}
 & &&  & $4$ & $4$  &  $120$&  \\
 \cmidrule{2-7}
&   \cite{hubara2018quantized} & $50$ & LSTM & FP & FP  &  {\bf 97}& \\
\cmidrule{5-7}
 &   &  &  & 4 & 4  &  $100$&  \\
 \cmidrule{2-7}
 & \cite{xualternating} & $30$ & LSTM & FP & FP &  {\bf 89.8}&  \\
  \cmidrule{5-7}
 &  &  &  & 2 & 2  &  $95.8$&  \\
   \cmidrule{4-7}
 &  &  & GRU & FP & FP  &  $92.5$&  \\
  \cmidrule{5-7}
 &  &  &  & 2 & 2  &  $101.2$&  \\
 \cmidrule{2-7}
 & \cite{kusupati2018fastgrnn}& $300$ & RNN & FP & FP  & $144.71$& \\
 \cmidrule{4-7}
 &  & & LSTM & FP & FP  & $117.4$&\\
 \cmidrule{4-7}
  & &  & skip-RNN & FP & FP  & {\bf 115.92}&  \\
 \cmidrule{4-7}
 & &  & FastGRNN & 8 & 16  & $116.11$&  \\
 \cmidrule{2-7}
 & \cite{wang2018hitnet}& $35$ & LSTM & FP & FP & {\bf 97.2}&   \\
 \cmidrule{5-7}
& &  & & t & t  & $110.3$& \\
 \cmidrule{4-7}
& &  & GRU & FP & FP  & $102.7$ & \\
 \cmidrule{5-7}
& &  &  & t & t  & $113.5$ & \\


\midrule
PTB & \cite{hubara2018quantized}& $50$ & RNN & FP & FP   & {\bf 1.05} & BPC \\
\cmidrule{5-7}
char.& &  & & 2  &  4  & 1.67& {\footnotesize(lower is better)}  \\ 
 \cmidrule{2-7}
 & \cite{Ott2016RecurrentNN}& $50$ & RNN & 1 & FP & {\bf 1.37} &\\
 \cmidrule{2-7}
 & \cite{ardakani2018learning}& $100$ & LSTM &FP & FP &  {\bf 1.39} & \\
\cmidrule{5-7}     
 & &  & & 1 & 12 &  1.43 & \\

\midrule
sequ. 
& \cite{ardakani2018learning}& $784$& LSTM &FP & FP & {\bf 98.9} & Accuracy  \\
\cmidrule{5-7}
MNIST& &  & & t & 12  & 98.8 &{\footnotesize (higher is better)}\\
\cmidrule{5-7}
& &  &  & 1 & 12 & 98.6& \\

 \cmidrule{2-7}
& \cite{kusupati2018fastgrnn}& $784$ &  LSTM & FP & FP &  97.8 &\\
\cmidrule{4-7}
& &  & skip-RNN & FP & FP &  {\bf 98.72} & \\
\cmidrule{4-7}
& &  &FastGRNN & 8 & 16 &  98.20 & \\
\cmidrule{4-7}
& &  &FastRNN & 8 & 16 &  96.44 & \\
 \cmidrule{2-7}
& \cite{foucault2025hadamrnn}& $784$ &  hadamRNN & 1/6 & FP &  96.9 &\\
\bottomrule
\end{tabular}
\end{table}

\myparagraph{Performance of RNNs: }
   Setting aside fastRNN temporarily, as indicated by the performances reported in  \Cref{results_biblio-tab} and \cite{gupta2022compression}, among the quantized RNNs, architectures follow the following general rule
\[
\mbox{LSTM} \gg \mbox{quantized LSTM} \qquad \mbox{and}\qquad \mbox{LSTM} \gg \mbox{GRU} \gg \mbox{quantized GRU}
\]
where \lq$\gg$\rq~means \lq has better performance than\rq. For this reason, we compare the QORNNs obtained by the proposed methods to the results of full-precision LSTM, which serves as an optimistic surrogate for all existing quantized LSTM and GRU architectures. We also compare our results to those of fastRNN and fastGRNN \citep{kusupati2018fastgrnn} and  \cite{foucault2025hadamrnn}. Beside HadamRNN, none of the existing quantized RNNs, LSTMs, or GRUs are able to solve the Copy-task with many time steps.

\subsection{Other quantized sequence models}\label{biblio-quantized-transformers-sec}

In addition to recurrent architectures, other families of sequence models, notably Transformers and state-space models (SSMs), have received increasing attention. These architectures are typically designed to benefit from large-scale training or pretraining datasets, which contrasts with the small, task-specific models considered in this work. Nevertheless, only a limited number of studies have addressed their quantization.

\myparagraph{Quantized transformers: } To the best of our knowledge \cite{shen2020q} was the first to address the quantization of weights and activations in Bidirectional Encoder Representations from Transformers (BERT), and as described in the recent survey \citep{tang2024survey}, many subsequent articles have followed suit.

Existing work on quantizing Transformer models such as BERT \citep{devlin2018bert} can be categorized according to the degree of quantization employed. Several methods concentrate on reducing weight precision to 8 bits \citep{zafrir2019q8bert,sun-etal-2020-mobilebert,stock2021training}, while others push this further to 4-bit representations \citep{shen2020q,zadeh2020gobo}. More extreme approaches investigate ternary-weight models \citep{zhang2020ternarybert} and binary-weight networks \citep{bai2021binarybert}. In addition, fully binarized variants of BERT, which also quantize activations, have been proposed in \citep{qin2022bibert,liu2022bit}.

Beyond BERT-scale models, recent work has also extended quantization techniques to very large language models with billions of parameters, notably through post-training or deployment-oriented approaches~\citep{yao2022zeroquant,frantar2023optq,xiao2023smoothquant,liu2023llm}.

These methods address a different operating regime from the one studied here, since their goal is to build extremely large models, while we design intrinsically small recurrent architectures.



\myparagraph{Quantized SSMs: }\label{biblio-quantized-SSMs-sec}
Recent research has increasingly focused on adapting quantization techniques to State Space Models (SSMs) to improve their deployability on resource-constrained platforms, revealing unique challenges due to their recurrent nature. Early work by Abreu et al.~ apply post-training quantization (PTQ) on the S5 model, \cite{abreu2024q}, demonstrated that while 8-bit quantization could be achieved with minimal accuracy loss (less than $1\%$ on several benchmarks), recurrent weights proved particularly sensitive, requiring higher precision than other components. Subsequent studies on the S4D architecture further highlighted the sensitivity of the state matrix and the internal state to PTQ. They showed that quantization-aware training (QAT) was necessary to recover performance at lower precisions~\cite{11345903}. The benefit in terms of model complexity is explored in \cite{siegel2026qs4d}. 

The article \cite{pierro2024mambaptq} first addresses these challenges in large, selective SSMs, like Mamba. Later Chiang et al.~proposed Quamba, in \cite{chiang2025quamba}. It is a PTQ recipe that handles activation outliers in the selective scan mechanism by suppressing input maxima and applying a Hadamard transform to create an outlier-free space for output quantization~\cite{chiang2025quamba}. This method successfully achieved 8-bit weight-activation quantization on a Mamba 2.8B model, reducing the model size to $2.76$GB, the latency by 1.72$\times$ on edge hardware with only a 0.9\% accuracy drop. An attempt to implement a PTQ of Mamba on FPGA is reported in \cite{wei2025lightmamba}.

As for quantized transformers, these networks still require at least several tens of megabytes of storage. This also contrasts sharply with the models considered in this article, whose storage requirements are significantly smaller (see \cref{glue_b_sec} and \cref{tab:glue_benchmark}).

\section{Preliminaries and notations}\label{background-sec}



In this section, we provide the main ideas and notations used 
on the RNN architecture, orthogonality, and quantization.

\subsection{Vanilla RNNs}\label{vRRNs-sec}
Vanilla RNNs define functions that take a time series as input and produce a vector (in the many-to-one case) or a time series (in the many-to-many case) as outputs. 

In order to define them, we consider positive integers $n_i$ and $T$, and an input time series $(x_t)_{t=1}^T \in (\RR^{n_i})^T$ with $T$ time-steps, made of $n_i$-dimensional data points. Denoting the output size $n_o\in\NN$, the output is either a vector in $\RR^{n_o}$ or a time series in $(\RR^{n_o})^T$.

The architecture of the RNN is defined by a hidden layer size $n_h\in\NN$, an activation function $\sigma$ and an output activation function $\sigma_o$. The parameters defining the vanilla RNN are $(W,U,V,b_o)$ for a recurrent weight matrix $W\in\RR^{n_h\times n_h}$, an input-to-hidden matrix $U\in\RR^{n_h\times n_i}$, a hidden-to-output matrix $V\in\RR^{n_o\times n_h}$, and bias $b_o\in\RR^{n_o}$. The hidden-state is initialized with 
$h_0=0$ 
and then computed using\footnote{Because biases in the recurrence unit are not central to the article’s presentation, we omit them for simplicity.}
\begin{equation} \label{eq:vrnn}
    h_t = \sigma \left( W h_{t-1} + U x_t 
    \right)\in\RR^{n_h},
\end{equation}
for $t\in\lb1,T\rb$.

In the many-to-one case, the output of the vanilla RNN is 
\[\sigma_o(Vh_{T} + b_o)\in\RR^{n_o}.
\]
In the many-to-many case, the output of the vanilla RNN is
\[\Bigl(\sigma_o(Vh_{t} + b_o)\Bigr)_{t\in\lb1,T\rb} \in (\RR^{n_o})^{T} .
\]
In all the experiments, $\sigma$ is either the Rectified Linear Unit (ReLU) or the modified ReLU  (modReLU)~\citep{helfrich2018orthogonal} activation functions, $\sigma_o$ is the identity function for regression tasks and the softmax function for classification tasks. 
The parameters $(W,U,V,b_o)$ are learned and $W$ is constrained to be quantized and approximately  orthogonal. 
The matrix $U$ is also quantized.

\subsection{Orthogonality}\label{ortho-sec}

\subsubsection{Orthogonal recurrent weight matrix}
The matrix $W\in\RR^{n_h\times n_h}$ is orthogonal if and only if
\[W^\mytop W = WW^\mytop = I,
\]
where $I$ denotes the identity matrix in $\RR^{n_h \times n_h}$ and $W^\mytop$ is the transpose of $W$.
This necessitates that the columns (respectively, rows) of the matrix possess a Euclidean norm of $1$, with the additional condition that any two distinct columns (respectively, rows) exhibit a scalar product of $0$. Among the various properties of orthogonal matrices, it is important to note that the singular values of orthogonal matrices are all equal to~$1$. Denoting $\sigma_{\min}(W)$ and $\sigma_{\max}(W)$ as the smallest and largest singular values of $W$, we have $\sigma_{\min}(W) = \sigma_{\max}(W) =1$. In other words, multiplication by an orthogonal matrix preserves norms. Orthogonal matrices constitute the Stiefel manifold  \citep{edelman1998geometry} that we denote $\STIEFEL{n_h}$.
\[\STIEFEL{n_h} = \{W \in\RR^{n_h\times n_h}~|~  W^\mytop W = WW^\mytop = I\}.
\]

The motivation behind constraining the recurrent weight matrix to be orthogonal is to mitigate instability and prevent issues such as vanishing or exploding gradients. This phenomenon has been  discussed in numerous articles, and we reiterate it for completeness in the next section.

\subsubsection{Memorization and stability, vanishing and exploding gradient}\label{ORNN-appendix}
For a large $T$, if the largest singular value $\sigma_{max}$ of the weight matrix $W$ is smaller than $1$, the initial entries of the input  $(x_t)_{t=1}^T$  cannot be effectively retained in the hidden state $h_T$. This prevents the consideration of long-term dependencies. Conversely, still considering large $T$, if the smallest singular value $\sigma_{min}$ of $W$ is greater than $1$, each multiplication by $W$ in \eqref{eq:vrnn} increases the magnitude of $h_t$, and the norm $\|h_t\|$ may tend towards infinity, leading to instability. For memorization and stability issues, it is desirable for the singular values of $W$ to remain close to $1$.

We arrive at the same conclusion when attempting to mitigate issues related to vanishing and exploding gradients. As indicated in \cite{Arjovsky2015UnitaryER} and echoed in subsequent literature on URNNs,  denoting $L$ the loss function, we find that:
\begin{eqnarray*} \label{eq:bptt}
    \frac{\partial L}{\partial h_t}
    &= & \frac{\partial L}{\partial h_T} \frac{\partial h_T}{\partial h_t} \\
    &= & \frac{\partial L}{\partial h_T} \prod_{i=t}^{T-1} \frac{\partial h_{i+1}}{\partial h_i} \\
    &= & \frac{\partial L}{\partial h_T} \prod_{i=t}^{T-1} D_i W^\mytop,
\end{eqnarray*}
where $D_i = \diag \bigl( \sigma' ( W h_{i-1} + U x_i ) \bigr)$ is the Jacobian\footnote{Since it is not central to our article, we assume that all the entries of  $W h_{i-1} + U x_i  $ are non-zero, ensuring that the Jacobian and $\sigma'$ are well-defined.} matrix of $\sigma$ evaluated at the pre-activation point and $W^\mytop$ is the transpose of $W$. If all the singular values of $W$ are less than $1$, those of $D_i W^\mytop$ are as well, causing the norm of $\frac{\partial L}{\partial h_t}$ to rapidly approach $0$ as $t$ decreases—resulting in the vanishing gradient problem. Conversely, if some singular values of $W$ are greater than $1$, depending on the activation patterns and $\frac{\partial L}{\partial h_T}$, the norm may explode, leading to the exploding gradient problem. To mitigate these phenomena, it is desirable for the singular values of $W$ to remain close to $1$. In other words, we aim for $W$ to be orthogonal or at least approximately orthogonal.

\subsubsection{Orthogonalizing a matrix}
In the models described in \cref{model-sec}, we consider two strategies that rigorously impose $W$ to be orthogonal. Both strategies establish a mapping from $\RR^{n_h\times n_h}$ to $\STIEFEL{n_h}$.

\begin{itemize}
    \item {\bf projUNN:} The first strategy employs the mapping $\PROJUNN$ as defined and implemented, referred to as projUNN-D, in \cite{kiani2022projunn}. This mapping computes the image $\PROJUNN(W)$ of matrix $W$ as the nearest orthogonal matrix in terms of the Frobenius norm. The implementation relies on a closed-form expression derived in \cite{keller1975closest}: $\PROJUNN(W)=W(W^\mytop W)^{-\frac{1}{2}}$. In the sequel, we use $\PROJUNN$ to implement a projected gradient descent algorithm solving a minimization problem involving an orthogonality constraint. 
    \item {\bf Björck:} 
    The second strategy  was introduced in \cite{59139e86-5891-3aa5-b8e6-33b9eeb82afa, Anil2018SortingOL} and applies a fixed and sufficiently large number of iterations of the following recursion
    \[
    A_{k+1}=\frac{3}{2}A_k-\frac{1}{2}A_kA_k'A_k \text{, initialized at } A_0 = \frac{1}{\sigma_{\text{max}}(W)} W,
    \]
     More details are given in %
     \cref{bjorck-appendix}. The resulting mapping from $\mathbb{R}^{n_h \times n_h}$ to the Stiefel manifold $\text{St}(n_h)$, denoted as $\BJORCK$, is surjective.
    Therefore, minimizing $L(\BJORCK(W))$ among unconstrained $W$ is equivalent to minimizing $L(W)$ among orthogonal $W$.
    Notice that standard backpropagation permits to compute $\left.\frac{\partial L\circ \BJORCK}{\partial W}\right|_{W} $.
\end{itemize}

%



\subsection{Quantization}\label{quant-sec}

We consider in this paper the most common scheme of quantization: a uniform quantization with a scaling parameter~\citep{rastegari2016xnor,gholami2022survey}.
For a quantization of bitwidth $k$, where $k \geq 2$, the
possible values are restricted to a set of size $2^k$, defined 
as follows:
\begin{equation}\label{Qk-eq}
\text{for a given }\Max > 0,\text{ } \calQ_k = \frac{\Max}{2^{k-1}} \left\lb -2^{k-1}, 2^{k-1} - 1 \right\rb.
\end{equation}
The set $\calQ_k$ evenly distributes values between $-\Max$ and $\Max - \frac{\Max}{2^{k-1}}$, with a quantization step of $\frac{\Max}{2^{k-1}}$.

For given $k$ and $\Max$, the quantizer $q_k$ maps every $W \in \RR^{n_h \times n_h}$ to the nearest element in $\calQ_k^{n_h \times n_h}$ based on the Frobenius norm. In other words, for every $(i,j) \in \lb1,n_h\rb^2$, the entry $(i,j)$ of the matrix $q_k(W)$, denoted $\bigl(q_k(W)\bigr)_{i,j}$, is the nearest element in $\calQ_k$ to $W_{i,j}$. 

When quantizing a matrix $W$, we take the value $\Max = \|W\|_{\max}$, where 
$\|W\|_{\max} = \max_{i,j} |W_{ij}|$. For ease of notation, we do not explicitly express the dependence on $\Max$. 
Note that, as is common practice~\citep{rastegari2016xnor}, when minimizing a function involving $q_k(W)$ with respect to $W$, we treat $\Max$ as a constant. Consequently, we do not backpropagate the gradient with respect to $\Max$.
To backpropagate through $q_k$, we employ the classical Straight-Through-Estimator (STE)
, described for completeness in \cref{ste-appendix}. An alternative could have been to use \cite{esser2020learned}.

\subsection{QORNNs are hard to train} \label{QORNN_hard-sec}
\begin{figure}[!t]
    \centering
    \begin{subfigure}[t]{0.49\textwidth}
    \centering
    \includegraphics[width=\linewidth]{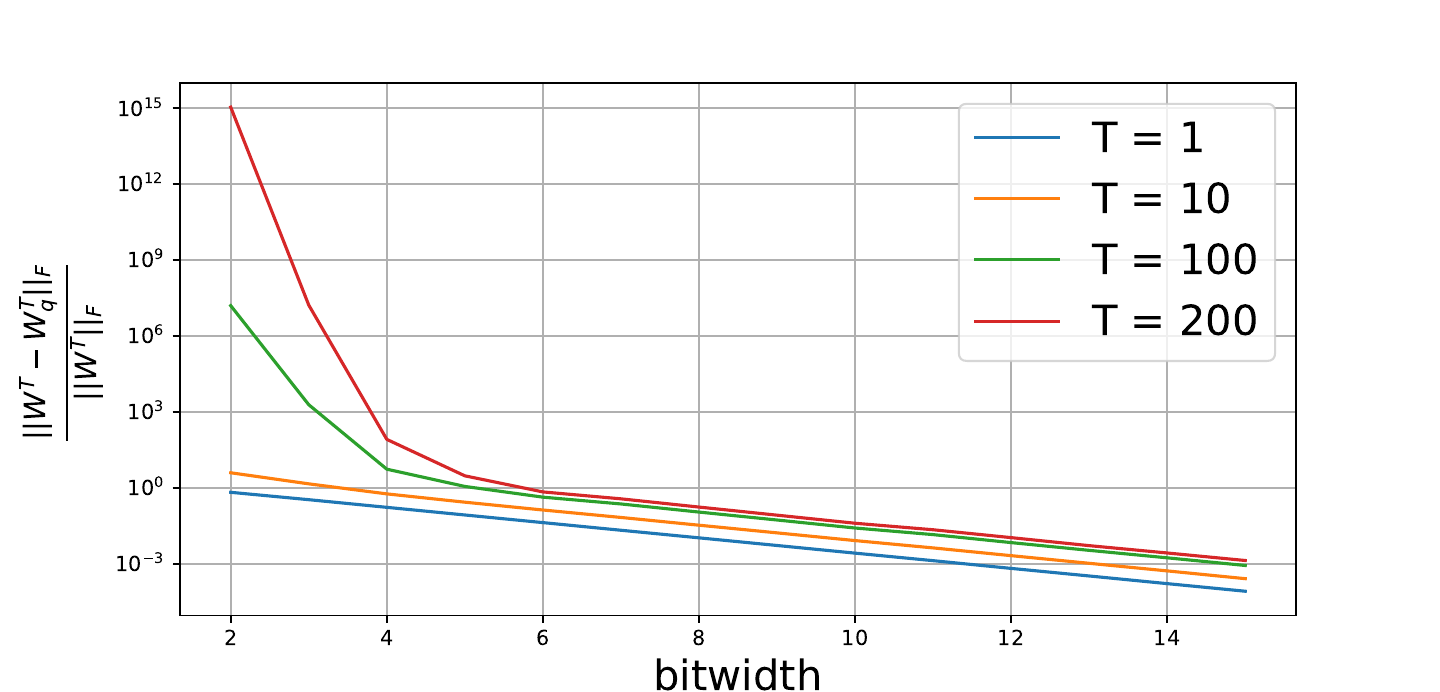}
        \caption{Illustration of compounding quantization error over $T$ recurrent applications of a quantized matrix.}
        \label{fig:sub1}
    \end{subfigure}
    \hfill
    \begin{subfigure}[t]{0.49\textwidth}
    \centering
        \includegraphics[width=\linewidth]{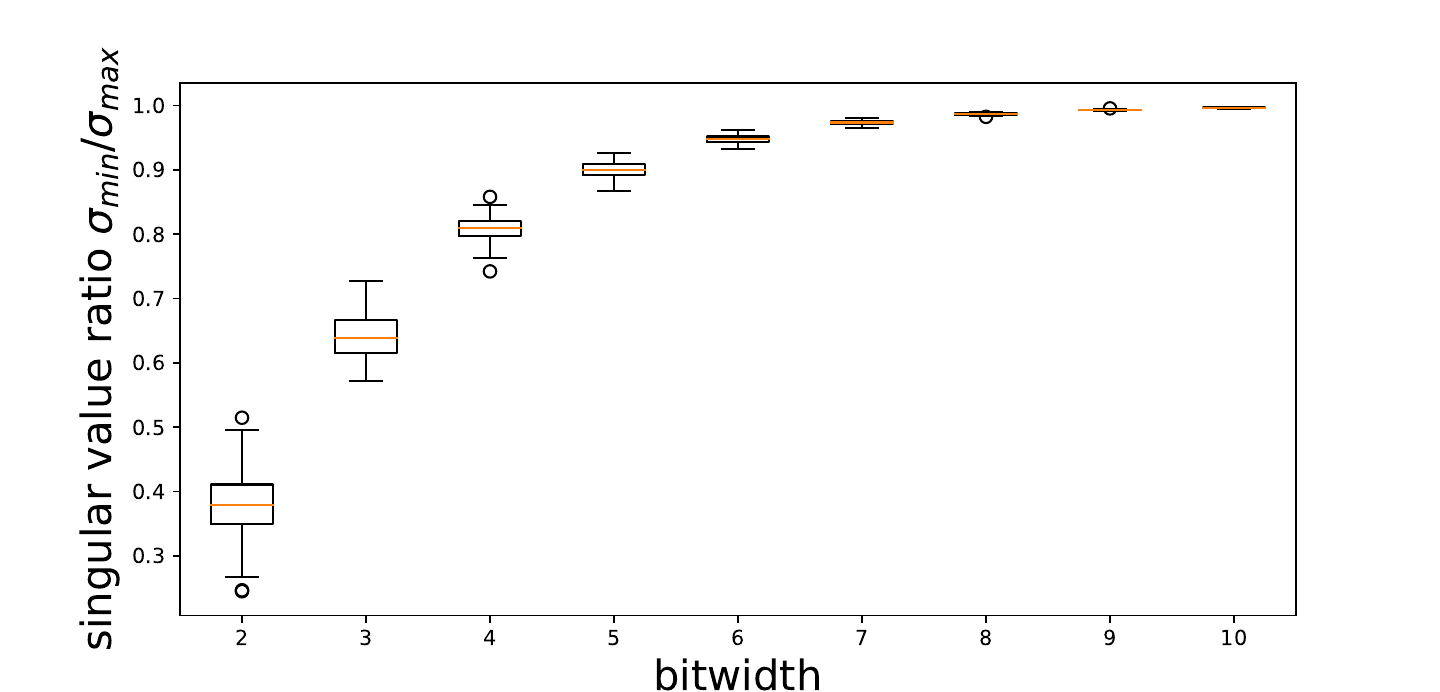}
 \caption{Deviation from orthogonality after quantization, measured by $\frac{\sigma_{min}}{\sigma_{max}}$, for $k\in\lb2,10\rb$. This ratio equals $1$ for orthogonal matrices; the further it deviates from $1$, the less orthogonal the matrix.}       
        \label{fig:sub2}
    \end{subfigure}

    \caption{Denote by $q_k$ the quantizer with bitwidth $k$ as defined in \cref{quant-sec}, $\sigma_{min}(q_k(W))$ and $\sigma_{max}(q_k(W))$ the smallest and largest singular values of the matrix $q_k(W)$ respectively for $W \in \RR^{200 \times 200}$ a uniformly sampled orthogonal matrix, and we write $W_q := q_k(W)$ for brevity.  (a) $\frac{\|W^T - (q_k(W))^T\|_F}{\|W^T\|_F}$ for various $k$ and powers, corresponding to the number of time steps, $T$. (b) Boxplots  for $1000$ random orthogonal matrices $W$ of the ratio $\sigma_{min}(q_k(W)) / \sigma_{max}(q_k(W))$ for various $k$.}

    \label{fig:probs_quant}
\end{figure}


    


\myparagraph{The instability problem:} We emphasize that quantizing vanilla RNN and ORNN is challenging. It is identified as a difficult unstable problem in \cite{Ott2016RecurrentNN,hou2017loss,hubara2018quantized}. The instability can be attributed to the following phenomena: 
\begin{itemize} 
\item The recurrent weight matrix is applied multiple times, rendering the network's output highly sensitive to even slight variations in the recurrent weight matrix. This also occurs during backpropagation.
\item The quantization of an orthogonal recurrent weight matrix generally results in a matrix that is not orthogonal. This, too, can contribute to the instability of the RNN.
\end{itemize}


Let us illustrate the first point.
We present in \cref{fig:probs_quant}-(a) the values of $\frac{\|W^T - q_k(W)^T\|_F}{\|W^T\|_F}$ for different bitwidths $k$ and various powers (analogous to the numbers of time steps) $T$ of the matrices\footnote{The analysis made in this paragraph does not take into account the effect of the activation function.
}.
Here, using the method described in \cite{mezzadri2007generate}, $W \in \RR^{200 \times 200}$ is a random matrix sampled according to a uniform distribution over orthogonal matrices of size $200\times 200$, $\|.\|_F$ represents the Frobenius norm,  $q_k$ is the quantization described in \cref{quant-sec}, and $T\in \{1, 10, 100, 200\}$. On \cref{fig:probs_quant}-(a), we see that $q_k(W)^T$ can be far from $W^T$, especially for small bitwidths $k$ and large time steps $T$.
As a result of this instability, the results of the forward pass using the quantized recurrent weights are far from the results for the full-precision recurrent weights, 
which may pose challenges in learning the quantized recurrent weights.

We illustrate the second point in \cref{fig:probs_quant}-(b): We present boxplots of the ratio $\sigma_{min}(q_k(W)) / \sigma_{max}(q_k(W))$, where $\sigma_{min}(q_k(W))$ and $\sigma_{max}(q_k(W))$ are respectively the smallest and largest singular values of $q_k(W)$. We uniformly sample $1000$ orthogonal matrices $W$ as described above, apply the quantizer $q_k$ to each of them for different bitwidths $k$, and compute the ratio $\sigma_{min}(q_k(W)) / \sigma_{max}(q_k(W))$. A ratio closer to $1$ indicates a higher level of orthogonality. The boxplots show that smaller bitwidths result in smaller expected ratios and more variable ratios.

\myparagraph{Evaluating the approximate orthogonality  of $q_k(W)$:}
 We obtain the following bounds for the orthogonality discrepancy of $W_q$.
\begin{proposition}
Assume $W$ is orthogonal and denote  $W_q=q_k(W)$. We have
    \begin{equation}\label{borneWq1}
\|W_q W_q^\mytop -I \|_F  \leq  2 \frac{n_h}{2^{k-1}}\|W\|_{\max} + \left(\frac{n_h}{2^{k-1}}\|W\|_{\max}\right)^2. 
\end{equation}
Similarly, we can derive bounds on  $\sigma_{\min}(W_q)$ and $\sigma_{\max}(W_q)$:
\begin{equation}\label{borneWq2}
1 - \frac{n_h}{2^{k-1}}\|W\|_{\max} \leq \sigma_{\min}(W_q) \quad\mbox{and}\quad  \sigma_{\max}(W_q) \leq 1 + \frac{n_h}{2^{k-1}} \|W\|_{\max}.
\end{equation}
We can complement these bounds using that, since $W$ is orthogonal, $\|W\|_{\max} \leq 1$.
\end{proposition}
\begin{proof}
Let us first prove \eqref{borneWq1}. Assume $W$ is orthogonal and denote $H = W_q - W$, where $W_q=q_k(W)$. Since $W$ is orthogonal, we have $\|WH^\mytop\|_F = \|H^\mytop\|_F= \|H\|_F$, and $W W^\mytop = I$. Using these equations, we obtain:
\begin{eqnarray*}
    \|W_q W_q^\mytop -I \|_F 
    & = & \|(W+H) (W+H)^\mytop - W W^\mytop \|_F \\
    & = & \|W H^\mytop + H W^\mytop + H H^\mytop \|_F\\
    & \leq & 2\|W H^\mytop\|_F + \|HH^\mytop\|_F \\
    & \leq & 2\|H\|_F + \|H\|^2_F.
    \end{eqnarray*}
Considering that, with the quantization scheme defined in \cref{quant-sec}, we have 
\begin{equation}\label{qervit}
    \|H\|_{\max}\leq \frac{\alpha}{2^{k-1}}= \frac{\|W\|_{\max} }{2^{k-1}}.
\end{equation}
We then deduce that $\|H\|_F \leq n_h \|H\|_{\max} \leq   \frac{n_h}{2^{k-1}} \|W\|_{\max}$. This leads to 
\[\|W_q W_q^\mytop -I \|_F \leq  2 \frac{n_h}{2^{k-1}} \|W\|_{\max} + \left(\frac{n_h}{2^{k-1}} \|W\|_{\max}\right)^2 
\]
and \eqref{borneWq1} holds.

We prove \eqref{borneWq2} similarly. We first remark that, using \eqref{qervit}, we also have $\sigma_{\max}(H)\leq n_h \|H\|_{\max} \leq \frac{n_h}{2^{k-1}}\|W\|_{\max}$. Using that $W$ is orthogonal, we obtain
\[
\sigma_{\min}(W_q)  \geq \sigma_{\min}(W) - \sigma_{\max}(H)  \geq 1 - \frac{n_h}{2^{k-1}}\|W\|_{\max}
\]
and
\[
 \sigma_{\max}(W_q) \leq \sigma_{\max}(W) +\sigma_{\max}(H)  \leq1 + \frac{n_h}{2^{k-1}} \|W\|_{\max}.
\]
We conclude that \eqref{borneWq2} holds.  
\end{proof}

\begin{table}[h!]
\centering
\caption{Values of $\frac{n_h}{2^{k-1}}$ for various $k$, with $n_h=256$ as used in the Copy-task experiment.}\label{tab_nh_k}
\begin{tabular}{c|ccccccccccccc}
\hline
 $k$                    & $2$   & $3$  & $4$  & $5$  & $6$ & $7$ & $8$ & $9$ & $10$  & $11$   & $12$ & $13$ & $14$\\
 $\frac{n_h}{2^{k-1}}$  & $128$ & $64$ & $32$ & $16$ & $8$ & $4$ & $2$ & $1$ & $0.5$ & $0.25$ & $0.125$ & $0.0625$ & $0.03125$\\    \hline
    \end{tabular}
\end{table}

\begin{figure}[!t]
    \centering
    \begin{subfigure}[b]{0.49\textwidth}
    \centering
    \includegraphics[width=\linewidth]{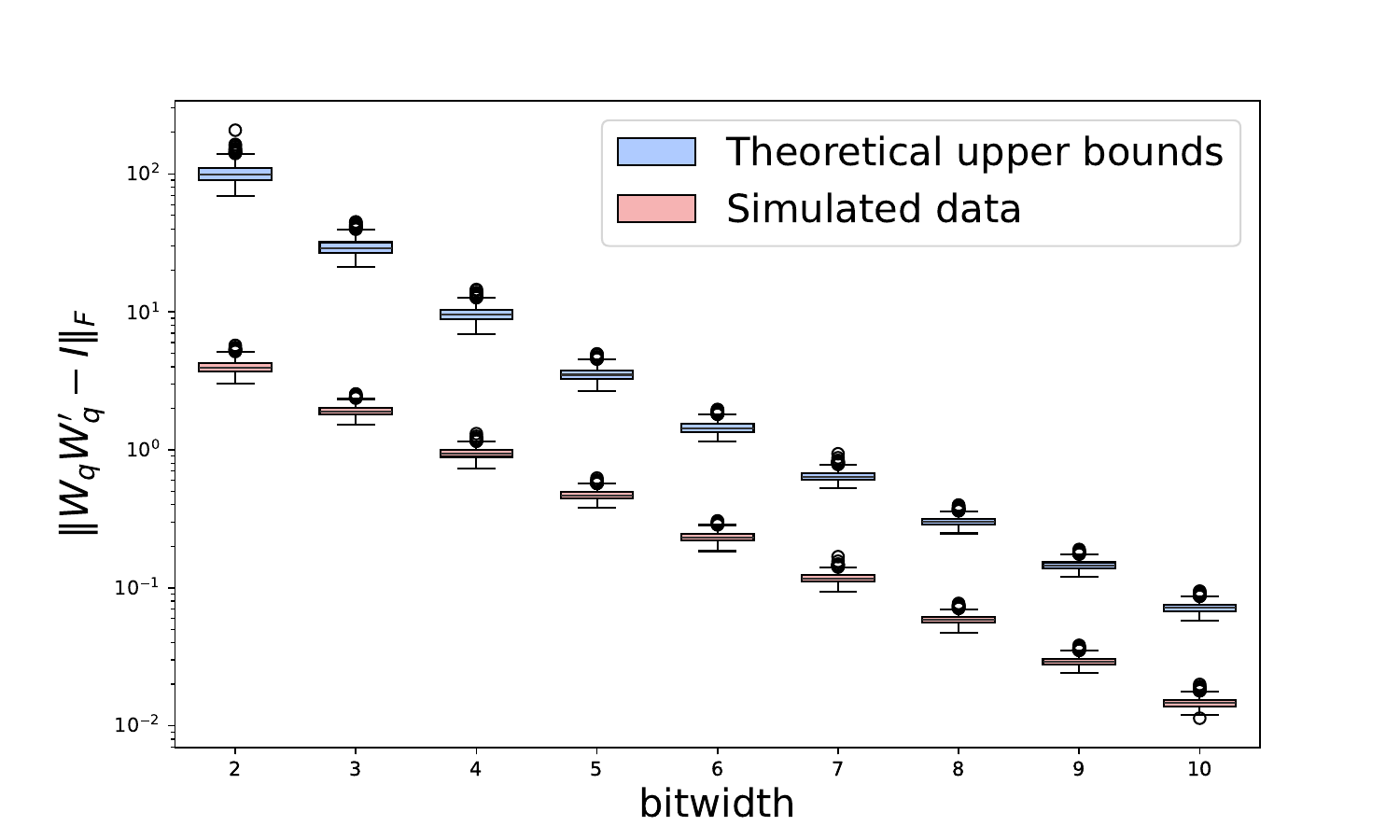}
        \caption{Empirical illustration of the upper-bound on $\|W_q W_q^\mytop - I\|_F$, in \eqref{borneWq1}.}
        \label{fig:sub3}
    \end{subfigure}
    \hfill
    \begin{subfigure}[b]{0.49\textwidth}
    \centering
        \includegraphics[width=\linewidth]{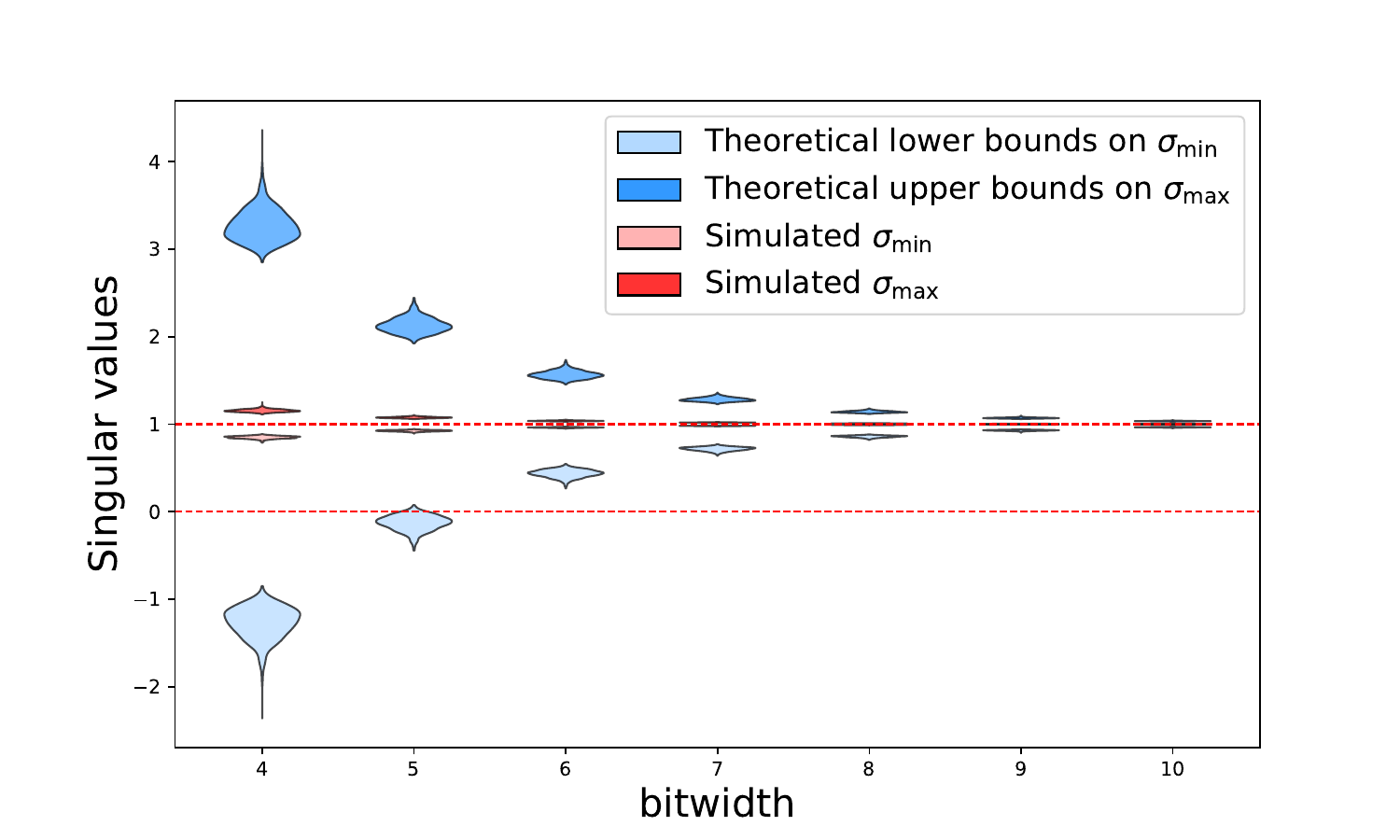}
        \caption{Empirical illustration of the bounds on $\sigma_{\min}$ and $\sigma_{\max}$, in~\eqref{borneWq2}.}
        \label{fig:sub4}
    \end{subfigure}
    

    \caption{Let $q_k$ denote the $k$-bit quantizer defined in \cref{quant-sec}. We sample $1000$ random orthogonal matrices $W \in \RR^{32 \times 32}$ and write $W_q := q_k(W)$. We report for several values of k (a) boxplots of the true $\|W_q W_q^{\mytop} - I\|_F$ and the corresponding upper bounds $32\|W\|_{\max}/2^{k-1}$ from \eqref{borneWq1} in log scale.  (b) violin plots of $\sigma_{\min}(W_q)$ and $\sigma_{\max}(W_q)$, along with the bounds $1 \pm 32\|W\|_{\max}/2^{k-1}$ from \eqref{borneWq2}.
    }
    \label{fig:bornes_quant}
\end{figure}

This permits obtaining guarantees of approximate orthogonality. However,  the theoretical guarantees are useful only when \( \frac{n_h \|W\|_{\max} }{2^{k-1}}\ll 1 \), a regime that is uncommon in practice-- see \Cref{tab_nh_k} and \cref{fig:bornes_quant}. The theoretical worst-case bounds are nonetheless pessimistic compared with the standard experimental results. Indeed, a graphical illustration of the bounds \eqref{borneWq1} and \eqref{borneWq2} is provided in \cref{fig:bornes_quant}. The figures are obtained by quantizing, for several bitwidth values \(k\), $1000$ random matrices sampled according to a uniform distribution over
orthogonal matrices of size 32 × 32, and then computing the quantities of interest for each quantized matrix. Note that the lower bound on \( \sigma_{\min} (W_q) \), shown in \Cref{fig:bornes_quant}(b), becomes non-positive (on average)---and therefore vacuous---for  \(k< 6\). Finally, as expected, we observe empirically that the singular values of the quantized matrix approach $1$ on average, as the bitwidth $k$ increases.

The experiments in \Cref{exp-sec} will show that the following proposed methods are effective in practice, with strategies that enforce both constraints during training, as detailed in the following section.

\section{Quantized RNNs with approximate orthogonality constraints}\label{model-sec}

In this section, we introduce the two strategies to build QORNN that are evaluated in \cref{exp-sec}.  The two strategies can be interpreted as different numerical schemes for solving the same highly non-convex optimization problem, as presented in the following subsections.

\subsection{Projected STE (STE-projUNN) }
A QAT strategy is applied to directly learn quantized weights with approximate orthogonality constraints $(q_k(W),q_k(U),V,b_o)$, for a given $k$, where $(W,U,V,b_o)$ are obtained using the {\it projected gradient descent algorithm} solving the following constrained optimization problem:
\begin{equation}\label{STE-projUNN-eq}
\left\{ \begin{array}{l}
\underset{(W, U, V, b_o)}{\min} ~ L(q_k(W), q_k(U), V, b_o) \\
W \mbox{ is orthogonal,}  
\end{array} \right.
\end{equation}
where $L$ is the learning objective. Notice that at each iterate of the algorithm, $W$ is constrained to be orthogonal. In the projected gradient descent algorithm, the gradients are computed using backpropagation and the STE, and the projections onto the Stiefel manifold are computed using $\PROJUNN$. \cref{alg:ste_projunn} details  each step of STE-projUNN\footnote{Quantization of $U$ matrix is omitted to simplify notations.}. We use the implementation of the reference code from  \cite{kiani2022projunn}  in the \href{https://github.com/facebookresearch/projUNN}{GitHub projUNN repository}.

\begin{figure}
    \begin{minipage}{0.49\textwidth}
        \begin{algorithm}[H]
            \caption{STE-ProjUNN algorithm}
            \label{alg:ste_projunn}
            \begin{algorithmic}[1]
    \Require Initial point: $W_0 \in \STIEFEL{n_h}$, learning rate: $\eta$, projection onto $\STIEFEL{n_h}$: $\PROJUNN$ 
    \State $i=0$
    \While{$x,y \in $ batches}
        \State $q_k(W_i)$ \Comment{Quantization} 
        \State $L(q_k(W_i), U, V, b_o,x,y)$\Comment{Forward pass}
        \State ${W'_i} = W_i - \eta.\nabla L$ \Comment{Weight update}
        \State $W_{i+1} = \PROJUNN({W'_i})$ \Comment{Project}
    \EndWhile
            \end{algorithmic}
        \end{algorithm}
    \end{minipage}
    \hfill
    \begin{minipage}{0.5\textwidth}
        \begin{algorithm}[H]
            \caption{STE-Björck algorithm}
            \label{alg:ste_bjorck}
            \begin{algorithmic}[1]
    \Require Initial point: $W_0\in\RR^{n_h\times n_h}$, learning rate: $\eta$, map onto $\STIEFEL{n_h}$: $\BJORCK$ 
    \State $i=0$
    \While{$x,y \in $ batches}
        \State $W'_i = \BJORCK(W_i)$\Comment{Map onto}
        \State $q_k(W'_i)$ \Comment{Quantization}
        \State $L(q_k(W'_i), U, V, b_o,x,y)$ \Comment{Forward pass}
        \State $W_{i+1} = W_i - \eta.\nabla L$  \Comment{Weight update}
    \EndWhile
            \end{algorithmic}
        \end{algorithm}
    \end{minipage}
\end{figure}

\subsection{STE with \texorpdfstring{$\BJORCK$}{Björck surjection} (STE-Björck)}\label{ste-bjork-sec}
A QAT strategy inspired by \cite{Anil2018SortingOL} is applied to directly learn quantized weights with approximate orthogonality constraints $(q_k(\BJORCK(W)),q_k(U),V,b_o)$, for a given $k$, where $(W,U,V,b_o)$ is obtained by a first-order algorithm solving the unconstrained optimization problem:
\[\underset{(W, U, V, b_o)}{\min} ~ L(q_k(\BJORCK(W)), q_k(U), V, b_o),
\]
where $L$ is the learning objective. The gradients are also computed using backpropagation and
the STE. Although $W$ is this time unconstrained, as already explained, since $\BJORCK$ is surjective onto the Stiefel manifold, solving this problem is equivalent to solving \eqref{STE-projUNN-eq}. However, the reformulation leads to a different algorithm, a priori facilitating the evolution of the recurrent weight matrix $W$. Besides, the recursive formulation of $\BJORCK$ given in \cref{ortho-sec}  makes it differentiable. \cref{alg:ste_bjorck} details each step of STE-Björck.

We use the opensource  \href{https://github.com/deel-ai/deel-torchlip}{GitHub Deel-Torchlip repository} from \cite{serrurier2021achieving} for $\BJORCK$ algorithm implementation.


\section{Activation quantization and inference complexity}\label{add-fullquant}
In this section, we compare the inference complexity of three  QORNNs variants: the full-precision \textit{Floating Point} RNN, the RNN whose weights have been quantized, called  \textit{Quantized Weights}, and the \textit{Fully Quantized} RNN. For the latter, we apply post-training quantization to the activations. Combined with the quantization-aware training of the weights described earlier, this enables fixed-point arithmetic computations, thereby yielding fully quantized RNNs. This post-training quantization strategy of activations  is detailed in \cref{add-fullquant-annexe}.

\cref{tab:computation_complexity-fm} gives the computational complexities for the matrix-vector multiplications appearing in the recurrent layer of the 
three variants: \textit{Floating Point}, \textit{Quantized Weights} and \textit{Fully Quantized} RNN.

For RNNs using quantized weights, i.e. for the third column of \cref{tab:computation_complexity-fm}, the complexity is derived from the binary decomposition of each quantized weight matrix. For instance, the recurrent matrix $W$ is represented as
\[q_{k}(W)= \frac{\alpha_W}{2^{k-1}}\left ( -2^{k-1}B_{k-1}+\sum_{i=0}^{k-2} 2^ i B_i \right )
\]
where, for all $i\in\lb 0,k-1\rb$, $B_i \in\{0,+1\}^{n_h\times n_h}$ is a binary matrix. 

For the \textit{Fully quantized} network described in this section, we obtain the complexities in the last column of \cref{tab:computation_complexity-fm}, depending on the quantization size of the input and recurrent features $k_i, k_a$ as described in \cref{add-fullquant-annexe}. 


\begin{table}[h!]
  \caption{Computational complexity for matrix-vector multiplications in the recurrent layer of the RNN. $\FP$ stands for in floating-point arithmetic, $\fp$ stands for fixed-point precision additions, $\fp_{l,l'}$ stands for fixed-precision multiplications between numbers coded using $l$ and $l'$ bits. We neglect the bit-shifts and the accesses to the look-up table.}
  \label{tab:computation_complexity-fm}
  \centering
  \begin{tabular}{lllll}
    \toprule
       Layer &Operation &   Full-precision   & Quantized weights & Fully Quantized \\
    \midrule
       Input matrix &Mult.&    $n_i. n_h~\FP$ &  $n_h~\FP$ & $n_i.n_h ~ \fp_{k,k_i}$\\
       &Add. &$n_i. n_h ~\FP$ & $ k.n_i.n_h~\FP$ &$n_i. n_h ~\fp$ \\
    \midrule
    Recurrent matrix & Mult.&   $n_h. n_h~\FP$ &  $n_h~\FP$ & $n_h.n_h ~ \fp_{k,k_a}$\\
      & Add. &$ n_h. n_h ~\FP$ & $ k.n_h.n_h~\FP$ &$n_h. n_h ~\fp
$\\
    \bottomrule
  \end{tabular}
\end{table}

Finally, in the experiments of \Cref{exp-sec}, for the Copy-task and PTB, since the inputs $x_t$ are one-hot encoded and therefore binary, the input layer can be computed using only $n_h$ multiplications and $(k_i-1).(n_i-1).n_h$ additions in floating point arithmetic, in the third column of \cref{tab:computation_complexity-fm},  and fixed-precision arithmetic in last columns of \cref{tab:computation_complexity-fm} respectively.

\section{Experiments}\label{exp-sec}

In this section, we present the results of the models described in \cref{model-sec} across several standard sequential tasks: the Copy-task in \cref{copy-sec}, the permuted and sequential pixel-by-pixel MNIST tasks (pMNIST and sMNIST, respectively)  in \cref{mnist-sec}, and the next character prediction task using the Penn TreeBank dataset in \cref{PTB-sec}. The first tasks are particularly challenging due to their reliance on long-term dependencies within the sequences, which makes them well-suited for ORNNs. Conversely, the Penn TreeBank task is a language model problem characterized by shorter-term dependencies. Similarly to the next character prediction task, sMNIST is known to be favorable to LSTMs. We also provide results on GLUE benchmark, in \Cref{glue_b_sec}, and IoT benchmarks in \Cref{har_dsa}. An additional task, the Adding task, is detailed in \cref{add-sec}.

To evaluate the performance of the QORNN, 
we also compare it with other full-precision RNNs
 or, when provided by other articles, their quantized counterparts and quantized transformers :
\begin{itemize}
\item LSTM \citep{hochreiter1997long} which also serves as an optimistic surrogate for all existing quantized models with the exception of FastRNN, see \cref{biblio-QRNN}.
\item ORNNs with the same hidden size as the quantized models, implemented using the projUNN-D  \citep{kiani2022projunn}, or Björck algorithms.
\item FastRNN \citep{kusupati2018fastgrnn}, either by using the figures from the original article or through additional experiments conducted with the reference code in floating-point.
\item Quantized transformers \citep{zafrir2019q8bert,zhang2020ternarybert,bai2021binarybert,liu2022bit,qin2022bibert} and HadamRNN \citep{foucault2025hadamrnn} are evaluated on the GLUE benchmark, using figures from the original articles.
\end{itemize}
In the subsequent results, instances where the RNNs did not outperform the naive baseline are denoted by {\it NC} (Not-Converged).

To verify whether the recurrence of learned QORNN could be fully quantized, we include additional results from a simple Post-Training-Quantization (PTQ) of the activations, as detailed in \Cref{add-fullquant}. Since quantizing the activations led to similar performance variations for STE-Björck and STE-ProjUNN, \cref{tab:performance_all} reports only the fully quantized results obtained with STE-Björck.

For each task, model sizes and hyperparameters were selected according to the loss value computed on a validation dataset.
A description of the problems, the training hyperparameters, additional results, and learning curves are provided in \cref{copy-appendix}, \cref{mnist-appendix}, \cref{PTB-appendix}, \cref{add-glue} and
\cref{add-sec}.




\begin{table}[t]
  \caption{{\bf Performance} of various models and bitwidths for weights and activations on the Copy-task, Sequential MNIST, Permuted MNIST, and Penn TreeBank Character Task (NC stands for \lq Not Converged\rq, NU stands for \lq Not useful because of other figures\rq, FP for \lq full-precision\rq). Source of figures: \blue{$^\dagger$} from ~\cite{ardakani2018learning}; \blue{$^*$} from~\cite{kiani2022projunn}
  ; \blue{$^\ddagger$} from \cite{kusupati2018fastgrnn}. 
   We analyze the stability of the learning procedure on PTB in \cref{tab:hyper}.
  }\label{tab:performance_all}
  \centering
  
  \begin{tabular}{l c c c c c c c c}  
    \toprule
    \multirow{2}{*}{Model}    & weight/activation & &\textbf{Copy-task} &\textbf{pMNIST} &\textbf{sMNIST} & \textbf{PTB}\\
    & bitwidth  & &cross-ent. & accuracy &accuracy & BPC \\
    \midrule
    \midrule
 
 \multirow{1}{*}{LSTM (\blue{$^*$}: \cite{kiani2022projunn},\blue{$^\dagger$}: \cite{ardakani2018learning})} & FP/FP & & NC & 92.00\blue{$^*$} & 98.90\blue{$^\dagger$}& 1.39\blue{$^\dagger$}\\
 \midrule
 \midrule
 \multirow{1}{*}{FastRNN (\blue{$^\ddagger$}: \cite{kusupati2018fastgrnn})} & FP/FP & & NC  & 90.83 & 96.44\blue{$^\ddagger$} & 1.455\\
 \multirow{1}{*}{FastGRNN} & FP/FP & & NC & 92.9 & NU  & 1.577\\
 \multirow{1}{*}{FastGRNN (\blue{$^\ddagger$}: \cite{kusupati2018fastgrnn})} & 8/16 & & NC  &  NU& 98.20\blue{$^\ddagger$} & NU\\
 \midrule
 \midrule
 \multirow{9}{*}{STE-Björck} & FP/FP &  & 6.4e-6   & 94.51 & 96.61 & 1.404 \\
     & 8/FP &  & 1.6e-5  & 94.64& 96.27 & 1.452  \\
     & 8/12 &  & 1.7e-5  & 94.76 & 96.20&  1.452 \\
     & 6/FP &  & 3.8e-5  & 93.93& 94.81 & 1.476 \\
     & 6/12 &  & 3.7e-5   & 93.94 & 94.74 & 1.476 \\
     & 5/FP &  & 2.5e-3  & 93.67& 87.75 & 1.490 \\
     & 5/12 &  & 2.5e-3  & 93.67& 87.70 & 1.490 \\
     & 4/FP &  & NC  & 92.36 & 73.84& 1.559  \\
     & 4/12 &  & NC  & 92.33& 73.38 &  1.559 \\
 \midrule
 \midrule
 ProjUNN (\blue{$^*$}: \cite{kiani2022projunn})
     & FP/FP &  & 1.1e-12  & 94.3\blue{$^*$}& 90.03 & 1.739 \\
     \multirow{3}{*}{STE-ProjUNN}  & 8/FP &   & 2.0e-10  & 91.27& 89.53 & 1.742 \\
     & 6/FP &  & 6.5e-5  &  90.73& 88.06 & 1.745 \\

    & 5/FP &  & 1.0e-3  & 90.89 & 87.42 & 1.753 \\
    \bottomrule
  \end{tabular}
\end{table}

\subsection{Copy-task}\label{copy-sec}

We build RNNs solving the Copy-task as described in \cite{wisdom2016fullcapacity}, based on the setup outlined by  \cite{hochreiter1997long}. The input is a sequence with $T = T_0 + 20$ time steps, where the initial 10 elements constitute a sequence for the network to memorize,  followed by a marker at $T_0+11$. The RNN's objective is to generate a sequence of the same number of time steps, with the last ten elements replicating the initial 10 elements of the input sequence. More details are given in \cref{copy-appendix}. This task is known to be a difficult long-term memory benchmark, that classical LSTMs struggle to solve \cite{Arjovsky2015UnitaryER,bai2018empirical,tallec2018can,kerg2019non,bai2019deep}, when $T_0$ is large.



The output activation $\sigma_o$ is the softmax, and the prediction error is measured using the average cross-entropy. The naive baseline has an expected cross-entropy of $\frac{10 \log 8}{T_0 + 20}$.

As in ~\cite{kiani2022projunn}, we conducted the experiments for $T_0=1000$ time steps, with ORNNs of size $n_h=256$. Details on the hyperparameters, learning curves, and results for $T_0=2000$ are provided in the \cref{copy-appendix}.

The fourth column of \cref{tab:performance_all} reports the performances for this task. As reported above, LSTM performances remains at the naive baseline level. FastRNN results were not documented in~\cite{kusupati2018fastgrnn}. Similarly to what was reported in \cite{kag2021time}, none of our experiments with this model achieved convergence, even with full precision weights. Conversely, both STE-projUNN and STE-Björck models converge when  $k \geq 5$. 
For $k=8$ the performance achieved by the QORNN nearly matches that of its floating-point counterpart. Our QORNN (configured with $k=5$ and $12$ bits activations) is the first reported RNN with a fully quantized recurrence capable of solving the Copy-task for $T_0=1000$. 
Additionally, the size of this QORNN is below $51$ kBytes, see \cref{tab:modelsize_all}.  

\subsection{Permuted and sequential pixel-by-pixel MNIST (pMNIST/sMNIST)}\label{mnist-sec}

\begingroup

\begin{table}[h!]
  \caption{\textbf{Model sizes} for Copy, MNIST, and Penn TreeBank tasks (Symbol $kP$ stands for  \lq kilo-parameters\rq; $kB$ for \lq kilo-Bytes\rq; NC stands for \lq Not Converged\rq, NU stands for \lq Not useful because of other figures\rq).
  Source of figures: $^\dagger$ from ~\cite{ardakani2018learning}; $^*$ from~\cite{kiani2022projunn}
  ;$^\ddagger$ from \cite{kusupati2018fastgrnn}}
  \label{tab:modelsize_all}
  \centering
  
  \begin{tabular}{l c c c c c c c c c c c c}  
    \toprule
    \multirow{2}{*}{Model}  & weight &  & \multicolumn{2}{c}{\textbf{Copy-task}} &  \multicolumn{2}{c}{\textbf{sMNIST}} &  \multicolumn{2}{c}{\textbf{PTB}} \\
     & bitwidth &  & $kP$ & size ($kB$)& $kP$ & size ($kB$)& $kP$ & size ($kB$) \\
    \midrule
    \midrule
 
 \multirow{2}{*}{LSTM} & FP &  & NC & NC & 41.4$^\dagger$ & 162$^\dagger$ &4300.8$^\dagger$ &16800$^\dagger$ \\
  & 2 &  & NC & NC & "  & 10$^\dagger$ &" & \textbf{525}$^\dagger$\\
 \midrule
 \midrule
 \multirow{1}{*}{FastRNN} & FP &  & NC & NC & 30.8 &  120.2 & 1151.0 & 4496 \\\midrule
 \multirow{1}{*}{FastGRNN} & FP &  & NC & NC & NU & NU  & 1151.0 & 4496\\\multirow{1}{*}{FastGRNN} & 8 &  & NC & NC & 6$^\ddagger$ & 6$^\ddagger$  & NU & NU\\
 \midrule
 \midrule
 & FP &  &  70.4 & 275  & 30.8 & 120.2 & 1151.0 & 4496\\
  STE-Björck      & 8 &  & " & 75.5 & "  & 35.0 & " & 1274  \\
  or    & 6 &  & " & 58.9 &  " & 27.9 & " & 1005.5 \\
  STE-projUNN  & 5 &  & "  & 50.6 & " & 24.4 & " & 871.2\\
    & 4 & & "  & NC & " & 20.8 & " & 737.0 \\
    \bottomrule
  \end{tabular}
\end{table}

\endgroup

These tasks are also challenging long-term memory problems. Here, data examples are the  $28 \times 28$ images from the MNIST dataset, where each image is flattened to a $784$-long sequence of $1$-dimensional pixels (normalized values in $[0,1]$). For pMNIST the pixels are randomly shuffled according to a fixed permutation.
The model has to predict the hand-written digit class (10 outputs).
Note that pMNIST  is a more challenging task for gated models such as LSTM,
and is in general not reported in the literature, see \cref{results_biblio-tab}. However it is a classical benchmark task for ORNN.

In this section, we fix $n_h = 170$ for all models. We use the ReLU activation with the STE-Björck strategies, and as recommended in \cite{kiani2022projunn} the modReLU activation with the STE-projUNN, see \cref{mnist-appendix} for details.

For sMNIST task, ORNNs in floating point achieve  performance comparable to FastRNN, albeit slightly inferior to that of gated models (LSTM and FastGRNN). The STE-projUNN strategy struggles to attain performance levels exceeding $90\%$. However, the STE-Björck model enables to achieve  an accuracy of $96.2\%$ with $k=8$ bitwidth weights quantization and a fully quantized recurrence. 

For pMNIST and STE-ProjUNN strategy, we could not replicate the projUNN performances reported in \cite{kiani2022projunn}. However, as shown in \cref{tab:performance_all}, applying STE-Björck, with $k=8$ bitwidth quantization for weight and 12 for activations, provides a QORNN with fully quantized recurrence that achieves results comparable to those reported in \cite{lezcano2019cheap, kiani2022projunn}, which currently represent the state-of-the-art performance on pMNIST with RNNs.
Interestingly, even for $k=4$, the  performance drop remains limited to $2.5\%$ with a network size with less than $21$ kBytes, see \cref{tab:modelsize_all}.

\subsection{Character level Penn TreeBank}\label{PTB-sec}

We present the results of QORNNs on a language modeling task. The Penn TreeBank dataset \cite{Taylor2003ThePT} (PTB), which comprises sentences of $150$ time steps ($T=150$), consisting of $50$ different characters ($n_i = 50$). 
The goal of the task is to predict the next character based on the preceding ones (further details can be found in \cref{PTB-appendix}). Similar to  sMNIST, this task is known to be favorable to LSTMs.
The purpose of this experiment is to offer a balanced assessment of performance and to evaluate the performance loss on a less favorable task. We also use PTB to perform further ablation studies in \cref{ablation-sec}. All experiments described in this section are for $n_h=1024$, except for the LSTM model, where the performance of \cite{ardakani2018learning} are reported, $1000$ cells. 
As reported in the literature, 
we use 
the Bit Per Character (BPC) measure .


The results for this task are reported in the last column of \cref{tab:performance_all} and \cref{results_biblio-tab}. The LSTM version with ternary weights proposed in \cite{ardakani2018learning}, see \cref{results_biblio-tab}, achieves the best results both in performance and size. For the QORNN, the STE-Björck strategy's performance is lower by 0.1 BPC with a network size of 872 kBytes.
\subsection{GLUE benchmark and comparison with transformers}\label{glue_b_sec}

After evaluating the proposed models on standard sequence-modeling benchmarks, we further assess their behavior in a more application-oriented natural language understanding setting. To this end, we consider the General Language Understanding Evaluation (GLUE) benchmark, which also enables a comparison with compact transformer-based models.

The General Language Understanding Evaluation (GLUE) is a widely used evaluation suite for natural language understanding, covering several sentence-level classification and inference tasks. State-of-the-art results on this benchmark are generally obtained with very large models trained on massive text corpora, most often pretrained transformer-based language models that are subsequently fine-tuned for each downstream task using a dedicated prediction head.

ORNNs, and consequently QORNNs, are not intended to be integrated into the standard large language model paradigm. Indeed, ORNN-based models do not appear in the GLUE leaderboard results. Nevertheless, we assess QORNNs on GLUE by training them from scratch on the training split of individual tasks. This provides a point of comparison with transformer-based approaches, while reflecting the intended lightweight training setting of these networks. More details on architecture and hyperparameters used are provided in \cref{add-glue}.

We considered two GLUE tasks. The first is called SST-2, it is a binary sentiment classification benchmark.
The second is QQP, the Quora Question Pairs task, where the objective is to determine whether two questions have the same meaning. In both cases, we followed the usual preprocessing pipeline: sentences were tokenized into sequences of integer indices and padded or truncated to a fixed length of 128 time steps. For paired-sentence inputs, we followed the reference implementation and concatenated the two sentences, inserting two separator tokens between them.

For each task, models were trained only on the official training set, without any additional data augmentation. The model selected for submission was the one reaching the best validation performance. Its predictions were then uploaded to the GLUE Benchmark server, and the corresponding official scores are reported in \cref{tab:glue_benchmark}. Since in previous experiments, STE-Björck surpasses STE-projUNN, we report only the former in this \cref{tab:glue_benchmark}.

Because QORNNs are optimized using only the data available for a single supervised task, increasing the model capacity quickly leads to overfitting. In particular, experiments using pretrained embeddings, such as the 768-dimensional first-layer representations from BERT, did not yield models that generalized well on the validation set. We therefore adopted a trainable embedding layer with dimension 256. 

On SST-2, the best result among the proposed models is obtained with the STE-Björck configuration using $n_i=256$, $n_h=64$, and $k=6$. This model reaches an accuracy of $83.4\%$ with a memory footprint of only 15 kBytes. This is within two percentage points of the smallest BiBERT model, which achieves $85.4\%$, while requiring approximately 293 times more storage. The smaller STE-Björck variants also remain competitive: with $k=4$, the model already reaches $82.4\%$ using only 10~kBytes, which is 0.5 percentage points higher than HadamRNN~\cite{foucault2025hadamrnn} for an increase of only 2~kBytes.

On QQP, the STE-Björck model with $n_i=256$, $n_h=512$, and $k=5$ achieves the best performance among our configurations, with an accuracy of $83.4\%$ and a size of 242 kBytes. This exceeds the smallest BiBERT model on QQP, which reaches $78.2\%$ with a size of 4.4 MB. Even compared with the intermediate 6.8 MB BiBERT model, which obtains $83.3\%$, the STE-Björck model with 5 bits quantization achieves a slightly higher accuracy while being about 29 times smaller. We note, however, that Block-HadamRNN~\cite{foucault2025hadamrnn} achieves the same performance as our $k=4$ model while using roughly one seventh of the memory, thanks to the ternary recurrent weight matrix.
The advantage of STE-Björck over Block-HadamRNN lies in the flexibility of the accuracy/size trade-off provided by the parameter \(k\).

\begin{table}[t]
    \small
    \centering
    \caption{Performance comparison of QORNN, and state-of-the-art models on two GLUE benchmark tasks: SST-2 and QQP. The last column indicates the model size in kilobytes (kB).
    }
    \label{tab:glue_benchmark}
    \begin{tabular}{lcc}
        \toprule
         Model &  Performance & Size\\
         &  & kBytes\\
         \midrule
         \midrule 
          \multicolumn{3}{c}{{\bf SST-2 task} ($T=128$, $n_o=1$, accuracy)} \\
          \midrule
          \midrule
        Q8BERT \citep{zafrir2019q8bert} &  92.24 & 110 000 \\
        Q-BERT \citep{shen2020q}&  92.08 & 48 100 \\
        TernaryBERT \citep{zhang2020ternarybert} &  92.8 & 18 000 \\
        MobileBERT \citep{sun-etal-2020-mobilebert} &  91.6 & 15 100 \\
        BinaryBERT \citep{bai2021binarybert} &  92.3 & 13 400 \\
        BiT \citep{liu2022bit} &  89.9 & 13 400 \\
        Bi{BERT} (\cite{qin2022bibert}) & 88.7 & 13 400\\
        Bi{BERT} \citep{qin2022bibert} & 87.9 & 6 800\\
        Bi{BERT} \citep{qin2022bibert} & 85.4 & 4 400\\
        HadamRNN \cite{foucault2025hadamrnn} &  81.9 & 8 \\
        STE-Björck (ours, $n_i=256$, $n_h=64$, $k=4$) &  82.4 & 10 \\
        STE-Björck (ours, $n_i=256$, $n_h=64$, $k=5$) &  83.3 & 13 \\
        STE-Björck (ours, $n_i=256$, $n_h=64$, $k=6$) &  83.4 & 15 \\
         \midrule
         \midrule 
          \multicolumn{3}{c}{{\bf QQP task} ($T=128$,  $n_o=1$, accuracy)} \\
          \midrule
          \midrule
        Q8BERT \citep{zafrir2019q8bert} &  87.96 & 110 000 \\
    TernaryBERT \citep{zhang2020ternarybert} &  88.8 & 18 000 \\
        BinaryBERT \citep{bai2021binarybert} &  88.9 & 13 400 \\
        BiT \citep{liu2022bit} &  85.4 & 13 400 \\
        Bi{BERT} \citep{qin2022bibert} & 84.8 & 13 400\\ 
        Bi{BERT} \citep{qin2022bibert} & 83.3& 6 800\\
        Bi{BERT} \citep{qin2022bibert} & 78.2 & 4 400\\
        Block-HadamRNN \cite{foucault2025hadamrnn} &  82.1 & 33 \\
        STE-Björck (ours, $n_i=256$, $n_h=512$, $k=3$) &  79.1 & 146 \\
        STE-Björck (ours, $n_i=256$, $n_h=512$, $k=4$) &  82.1 & 194\\
        STE-Björck (ours, $n_i=256$, $n_h=512$, $k=5$) &  83.4 & 242 \\
         \bottomrule
    \end{tabular}
\end{table}

\subsection{Two IoT benchmarks: HAR-2 and DSA-19}
\label{har_dsa}

QORNNs offers a more compact model representation and reduced computational overhead compared to traditional RNN architectures, enabling their deployment on resource-constrained devices. Such devices are typically employed in \textit{Internet of Things} (IoT) applications.

We evaluate QORNN on two small-scale, short-term memory IoT datasets: the \textit{Human Activity Recognition} (HAR-2) and  the \textit{Daily and Sport Activity} (DSA-19). These are two human motion datasets in which physical activity is recorded using body-worn sensors. Each data point corresponds to a sequence extracted from sensor signals, and the task is to predict the associated human activity or posture.

These two benchmarks were notably investigated in \citep{kusupati2018fastgrnn, foucault2025hadamrnn}, where performance and complexity were assessed. We integrate their results alongside those of QORNN's in \cref{tab:IoT}.

Following \citep{kusupati2018fastgrnn}, the six different position labels in HAR-2 (\textit{Sitting}, \textit{Laying}, \textit{Walking upstairs}, \textit{Standing}, \textit{Walking}, \textit{Walking downstairs}) were grouped into two categories: (i) Sitting, Laying, Walking upstairs; and (ii) Standing, Walking, Walking downstairs, thereby casting the problem as a binary classification task. We used $10\%$ of the training set for validation, and the model achieving the best score on the validation set was selected for submission. For both HAR-2 and DSA-19, each dataset's features were standardized to have zero mean and unit variance. More details on architecture and hyperparameters used are displayed in \cref{app:iot}.

On DSA-19, the best performance is obtained with the STE-Björck configuration. Our QORNN reaches $91.7\%$ accuracy, using a hidden dimension $n_h = 32$ and a bitwidth $k=2$, outperforming the state of the art (FastGRNN), by approximately $6\%$. Furthermore, the resulting model requires only $3.2$ kBytes of memory, compared to $22$ kBytes for the reported FastGRNN model, corresponding roughly to a $6.9 \times$ model size reduction. We also report the performances of QORNNs with bitwidths $k = 3, 5, 7$, and observe, as expected, that the accuracy improves with $k$.

On HAR-2, we also report results using the STE-Björck configuration, with a hidden dimension $n_h=32$. QORNN achieves higher accuracy than HadamRNN with a bitwidth $k=2$, which corresponds to a model size of $0.6$ kByte compared to $5.8$ kBytes for HadamRNN. With an $8$-bits precision, QORNN attains an accuracy of $93.17\%$, which is within $2.5$ points of both FastGRNN and Block-HadamRNN, while these models are respectively $2 \times$ and approximately $4.7 \times$ larger.

Altogether, these experiments show that on these IoT tasks, QORNN either improves upon the state of the art or enables a new performance–size trade-off.

\begin{table}[h!]
    \small
    \centering
    \caption{Performance comparison of QORNN, and state-of-the-art models on two IoT benchmarks: HAR-2 and DSA-19. 
    }
    \label{tab:IoT}
    \begin{tabular}{lcc}
        \toprule
         Model &  Performance & Size\\
         &  & kBytes\\
         \midrule
         \midrule
         \multicolumn{3}{c}{{\bf DSA-19 task} ($T=125$, $n_i = 45$, $n_o=19$, accuracy)} \\
          \midrule
          \midrule
        FastGRNN \citep{kusupati2018fastgrnn}&  85.67 & 22 \\
        HadamRNN \citep{foucault2025hadamrnn} & 79.12 & 34.1 \\
        Block-HadamRNN \citep{foucault2025hadamrnn} &  85.26 & 26.1 \\
        STE-Björck (ours, $n_h=32$, $k=2$) & 91.71 & 3.2 \\
        STE-Björck (ours, $n_h=32$, $k=3$) & 93.51 & 3.5 \\
        STE-Björck (ours, $n_h=32$, $k=5$) & 95.81 & 4.1 \\
        STE-Björck (ours, $n_h=32$, $k=7$) & 98.0 & 4.7 \\
         \midrule
         \midrule
          \multicolumn{3}{c}{{\bf HAR-2 task} ($T=128$, $n_i = 9$, $n_o=1$, accuracy)} \\
          \midrule
          \midrule
        FastGRNN \citep{kusupati2018fastgrnn}& 95.59 & 3 \\
        HadamRNN \citep{foucault2025hadamrnn} &  84.45 & 5.8 \\
        Block-HadamRNN \citep{foucault2025hadamrnn} & 95.64 & 7.1 \\
        STE-Björck (ours, $n_h=32$, $k=2$) & 89.95 & 0.6 \\
        STE-Björck (ours, $n_h=32$, $k=3$) & 88.62 & 0.7 \\
        STE-Björck (ours, $n_h=32$, $k=5$) & 90.20 & 1.1 \\
        STE-Björck (ours, $n_h=32$, $k=8$) & 93.17 & 1.5 \\
     \bottomrule
    \end{tabular}
\end{table}

\subsection{Ablation study}\label{ablation-sec}
In this section, we evaluate the influence QAT and projection ($\PROJUNN$ or $\BJORCK$) in the proposed solutions.
We 
study 
two additional strategies relaxing one of these constraints:
\begin{itemize}
    \item PTQ strategy, detailed in \Cref{ptq-sec},  corresponds to training a full-precision ORNN without any QAT constraint, and applying Post-Training Quantization on the learnt weights for all values of $k$.
    \item STE-pen strategy, detailed in \Cref{pen-sec}, imposes soft-orthogonality using a regularization term of the form $\lambda \|q_k(W)(q_k(W))^\mytop - I \|_F$, where $\lambda$ is a trade-off parameter. The quantized model is optimized using the STE and does not use any kind of projection.
\end{itemize} 

The results are in \Cref{ablation-results-sec}.

\subsubsection{Post-Training Quantization (PTQ)}\label{ptq-sec}

For any value of $k$, the weights with approximate orthogonality constraints, and quantized, after the training phase, using $k$ bits, are $(q_k(W),q_k(U),V,b_o)$, where $(W,U,V,b_o)$ is
the full-precision parameters obtained using the {\it projUNN-D} algorithm \cite{kiani2022projunn} for solving
\[\left\{ \begin{array}{l}
\underset{(W, U, V, b_o)}{\min} ~ L(W, U, V, b_o)   \\
W \mbox{ is orthogonal,}  
\end{array} \right.
\]
where $L$ is the learning objective.

\subsubsection{Penalized STE (STE-pen)}\label{pen-sec}
A Quantization-Aware-Training (QAT) strategy is applied to directly learn quantized weights $(q_k(W),q_k(U),V,b_o)$, with approximate orthogonality constraints, for a given $k$. The weights $(W,U,V,b_o)$ are obtained using an implementation of the Straight-Through Estimator (STE) to solve the following optimization problem:
\begin{equation}
\label{eq:ste_pen_obj}
\underset{(W, U, V, b_o)}{\min} ~ L(q_k(W), q_k(U), V, b_o) + \lambda R(q_k(W)).
\end{equation}
Here, $L$ represents the learning objective, $R$ is the regularization term enforcing orthogonality as defined by
\[
R(W) = \|W W^\mytop - I\|_F^2,
\]
and $\lambda$ is a parameter that balances the trade-off between minimizing $L$ and $R$.

For the experiment reported in \cref{ablation-results-sec}, the regularization parameter $\lambda$, which governs the trade-off between optimizing the learning objective and the regularizer  is set to $1e-1$.

\subsubsection{Ablation results}\label{ablation-results-sec}


\begin{table}[h!]
  \caption{Ablation study: comparison of STE-Björck, PTQ, STE-pen on permuted MNIST task}
  \label{tab:ablation}
  \centering
  \begin{tabular}{l c c c c c}
    \toprule
   Task      &  bitwidth $k$ & STE-Björck &    PTQ &  STE-pen   \\
    \midrule
 \multirow{3}{*}{pMNIST}   & 4 & 92.36 &  13.3 & 44.9
 \\
     &  6 & 93.93 &  42.2 & 75.4
 \\
     &  8 & 94.64 &  90.2 & 71.0 
 \\
    \bottomrule
  \end{tabular}
\end{table}

\cref{tab:ablation} provides a comparison on permuted MNIST. PTQ induces a large drop in performance for bitwidths $k<8$. This illustrates the difficulty of the problem. Moreover, penalization fails to learn effectively across all bitwidths and learning is unstable. This confirms that both QAT and projections are essential for learning QORNN models. 


\subsection{Hyperparameter exploration}
\subsubsection{Hidden layer size versus bitwidth}
In this section, we complement the experiments whose results are reported in \cref{tab:performance_all} with the results for other hidden layer size. The results show that the hidden layer size and bitwidth can be adjusted to achieve different performance/size trade-offs.

\myparagraph{On the Copy-task:} We present the results for the Copy-task in \cref{tab:performance_copy_190} with $T_0 = 1000$ time steps and $n_h = 190$. The conclusions drawn are similar to those depicted in \cref{tab:performance_all}, where $T_0=1000$ and $n_h=256$. The main difference between the results in \cref{tab:performance_copy_190} for $n_h = 190$ and \cref{tab:performance_all} for $n_h=256$ lies in the fact that a smaller value of $n_h$ allows achieving comparable performance but with a larger bitwidth $k$. There appears to be a trade-off between hidden size and bitwidth. Ideally, the trade-off should be tuned to achieve a reduction in the network's size.

\begin{table}[h!]
  \caption{Performance for STE-Björck and STE-ProjUNN for the Copy-task for $T_0=1000$ and $n_h=190$ for several weight bitwidths.
  }\label{tab:performance_copy_190}
  \centering
  
  \begin{tabular}{l c c c c c c c }  
    \toprule
    \multirow{2}{*}{Model}    &  \multirow{2}{*}{$T_0$}  & \multirow{2}{*}{$n_h$} &&\multicolumn{4}{c}{weight} \\
    &  &  & & \multicolumn{4}{c}{bitwidth} \\
    \midrule
    &  &  & & FP & 5 & 6 & 8 \\
 \midrule
 \midrule
 \multirow{1}{*}{STE-Björck} & 1000 & 190  &  & 8.1e-6 &1.4e-2 & 1.0e-3& 3.2e-05\\
 \midrule
 STE-ProjUNN
     & 1000 & 190 &   & 6.0e-10 & 2.9e-3 & 6.4e-4 & 2.4e-08  \\
    \bottomrule
  \end{tabular}
\end{table}

\myparagraph{On pMNIST:} We present the results for pMNIST in \cref{tab:pmnist360} with a larger hidden size, $n_h = 360$. The qualitative conclusions drawn are similar to those depicted in \cref{tab:performance_all} for $n_h = 170$. Note that STE-Björck with $n_h=360$ achieves an accuracy higher than $93\%$ even for $3-bits$ quantization but with a model parameter size of $62$ kBytes.

\begin{table}[h!]
  \caption{Performance for STE-Björck and STE-ProjUNN  on the pMNIST $n_h=360$ for several weight bitwidths
  }\label{tab:pmnist360}
  \centering
  
  \begin{tabular}{l c c c c c c c }  
    \toprule
    \multirow{2}{*}{Model}    & \multirow{2}{*}{$n_h$} & &\multicolumn{5}{c}{weight} \\
    &  &   &\multicolumn{5}{c}{bitwidth} \\
    \midrule
    &  &   & FP & 3 & 5 & 6 & 8 \\
 \midrule
 \midrule
 \multirow{1}{*}{STE-Björck}  & 360  &  & 95.43  & 93.32 & 95.71 & 95.51 & 95.20\\
 \midrule
 STE-ProjUNN
      & 360 &   & 93.28 & 44.29 & 92.17 & 93.06 & 93.10\\
    \bottomrule
  \end{tabular}
\end{table}

\myparagraph{On Penn TreeBank:} 
\Cref{tab:PTB2048} presents supplementary results, for a larger hidden size of $n_h = 2048$, on the Penn TreeBank dataset. Most results are similar to those reported in \cref{tab:performance_all} and obtained with $n_h = 1024$. While STE-ProjUNN  strategy achieves better results  than those with $n_h = 1024$, they are still inferior to the one obtained by STE-Björck strategy. Due to the larger hidden layer size, the results of STE-Björck are improved, but at the expense of an increased network size.

\begin{table}[h!]
  \caption{Performance for STE-Björck and STE-ProjUNN on  Penn TreeBank with $n_h=2048$ and several weight bitwidths
  }\label{tab:PTB2048}
  \centering
  
  \begin{tabular}{l c c c c c c c }  
    \toprule
    \multirow{2}{*}{Model}    & \multirow{2}{*}{$n_h$} & &\multicolumn{4}{c}{weight} \\
    &  &   &\multicolumn{5}{c}{bitwidth} \\
    \midrule
    &  &   & FP & 4 & 5 & 6 & 8 \\
 \midrule
 \midrule
 \multirow{1}{*}{STE-Björck}  & 2048  &  & 1.45  & 1.53 & 1.45 & 1.43 & 1.45\\
 \midrule
 STE-ProjUNN
      & 2048 &   &1.60 & NC & 1.79 & 1.66 &1.60\\
    \bottomrule
  \end{tabular}
\end{table}

\subsubsection{Stability with regard to optimization hyperparameters}

\cref{tab:hyper} presents the influence of the optimization hyperparameters on the performance of the PTB task for the bitwidth $k=5$. Experiments were done for other tasks and other bitwidths with the same conclusions and are not reported.

On the top of \cref{tab:hyper}, we see that the initial value of the learning-rate impacts the convergence, as it is often the case: very small learning-rates tend to evolve slowly, requiring more epochs, while excessively large rates fail to learn. The range of acceptable learning-rates is reasonably large.

In the middle of \cref{tab:hyper}, we see that batch size influences performance, with larger batch sizes generally leading to worse outcomes. This effect may be related to the number of update steps during training and could be mitigated by increasing the number of epochs. 

To obtain the figures on the bottom of \cref{tab:hyper}, we run STE-Björck $5$ times, starting each time from a different random initialization. We report the minimum, median, and maximum values of the BPC.
We observe a variation of amplitude $0.01$ BPC depending on the random initialization changes. We also observed, but do not report here, that, as expected, the greater the bitwidth, the smaller the variance.

\begin{table}[h!]
  \caption{Influence of the optimization hyperparameters (learning rate, batch size, and initialization) on the performance of STE-Björck at bitwidth $k=5$ on the Penn Treebank dataset.
}
  \label{tab:hyper}
  \centering
  \begin{tabular}{l c c c c }
    \toprule
Learning Rate   & 1e-4 & 1e-3 & 1e-2 & 1e-1 \\ 
Perf. (BPC) &1.803 &1.490 & 1.506 & 2.087  \\\midrule
Batch size    & 64 & 128 & 256& 512 \\
Perf. (BPC) & 1.482 & 1.490 & 1.510 & 1.570 \\\midrule
Random init. & min BPC & median BPC & max BPC& 
 \\
Perf. (BPC) &  1.484& 1.490& 1.494 & \\
    \bottomrule
  \end{tabular}
\end{table}

\subsection{Complements on training computation time}\label{app-compute}

\cref{tab:compute} presents the training computation time per epoch for each task and each model. Experiments were done on a NVIDIA GeForce RTX 3080 GPU.

\begin{table}[h!]
    \caption{Training computation time for different models and several tasks}
    \label{tab:compute}
  \centering
  \begin{tabular}{l l c c c  c  c  c }  
    \toprule
    \multirow{2}{*}{Task }    &\multirow{2}{*}{Model}    & \multirow{2}{*}{$n_h$} &\multirow{2}{*}{$T$} & epoch compute \\
    & &  &   & time (minutes) \\
    \midrule
 \midrule
 \midrule
 \multirow{2}{*}{\textbf{Copy-task}}& STE-Björck
      & 256 & 1020  & 38 \\
 & STE-ProjUNN
      & 256 & 1020  & 39 \\
 \midrule
 \midrule
 \multirow{4}{*}{\textbf{pMNIST}}  & FastRNN
      & 170 & 784  & 3.8 \\
 & FastGRNN
      &170& 784  & 4.2 \\
 & STE-Björck
      & 170 & 784  & 2.2\\
 & STE-ProjUNN
      & 170 & 784  & 4.2 \\
 \midrule
 \midrule
 \multirow{4}{*}{\textbf{PTB}}& FastRNN
      & 1024 & 150  & 0.57 \\
 & FastGRNN
      & 1024  & 150  & 0.86 \\
 & STE-Björck
      &  1024  & 150  & 1.07\\
 & STE-ProjUNN
      & 1024  & 150 & 0.37 \\
    \bottomrule
  \end{tabular}

\end{table}

\section{Conclusion and perspectives}

In this article, we propose and study two algorithms to construct Quantized approximately Orthogonal RNNs (QORNNs). They enjoy the benefits of ORNNs and work when ORNNs do. In particular, they manage to solve the Copy-task for $T_0=1000$ and $2000$, which existing quantized RNNs were unable to do. The model called STE-Björck also surpasses existing quantized models on tasks requiring long-term dependencies, such as permuted MNIST as well as IoT tasks such as DSA-19. However, they are less effective on tasks, such as sequential MNIST and PTB, where LSTMs and other gated RNNs outperform orthogonal RNNs. Their flexibility always makes it possible to achieve performance/model size trade-offs that are not feasible with other existing methods.
Additional experiments on GLUE and IoT benchmarks further show that QORNNs remain competitive in more application-oriented settings, while preserving very small memory footprints.

We demonstrate that combining orthogonalization and quantization-aware training is crucial for effectively training QORNN. In most experiments, this combination is more efficient when using the Björck orthogonalization method.


Future work on QORNNs could focus: 1/ on implementing QORNNs on dedicated hardware, 2/ on developing learning approaches better taking into account the activation quantization. It is also very much relevant to work on quantizing other models such as SSM to target longer dependencies such as the Long Range Arena benchmarks described in \cite{tay2020long}.




\bmhead{Acknowledgements}
This work has benefited from the AI Interdisciplinary Institute ANITI, which is funded by the French ``Investing for the Future – PIA3'' program under the Grant agreement ANR-19-P3IA-0004. The authors gratefully acknowledge the support of the DEEL project\footnote{\url{https://www.deel.ai/}}. A. Foucault was supported by \lq R\'egion  Occitanie, France\rq, which provided a PhD grant. Part of this work was performed using HPC resources from CALMIP (Grant 2024-P22034).  The work also benefited from the support of the PEPR IA project MadLearning\footnote{\href{https://www.pepr-ia.fr/projet/madlearning-2/}{https://www.pepr-ia.fr/projet/madlearning-2/}}, under grant agreement ANR-25-PEIA-0002.




\begin{appendices}

\section{The Björck algorithm}\label{bjorck-appendix}
Björck algorithm \cite{59139e86-5891-3aa5-b8e6-33b9eeb82afa} aims to minimize the regularization term 
\[R(W) = \|W W^\mytop - I\|_F^2,
\]
As described in \cite{Anil2018SortingOL}, the initialization is done with the matrix $A_0 = \frac{1}{\sigma_{\text{max}}(W)} W$, where $\sigma_{\text{max}}(W)$ is the largest singular value of $W$, computed using the power iteration for a fixed number of iterations. Then it applies a fixed and sufficiently large number of iterations, in practice $15$, of the following operation:   
\[
A_{k+1}=A_k\left(I+\sum_{i=1}^p(-1)^p \binom{-\frac{1}{2}}{p} Q_k^p\right)
\]
where $Q_k=I-A'_kA_k$ and $\binom{z}{p} = \frac{1}{p!} \prod_{i=0}^{p-1} (z-i)$. As described in \cite{Anil2018SortingOL}, we take $p=1$. In this case, the Björck algorithm corresponds to several iterations of the gradient descent algorithm minimizing $R$:
\[A_{k+1}=A_k-\frac{1}{2}A_k(A_k'A_k-I) = \frac{3}{2}A_k-\frac{1}{2}A_kA_k'A_k,
\]
initialized at $A_0 = \frac{1}{\sigma_{\text{max}}(W)} W$. 

We compute $\left.\frac{\partial \BJORCK}{\partial W}\right|_{W}$ using standard backpropagation but treat $\sigma_{\text{max}}(W)$ as a constant.
\section{The straight-through-estimator}\label{ste-appendix}

Considering $\alpha$ as fixed, the mapping $W \longmapsto q_k(W)$ is piecewise constant. Its gradient at $W$, denoted as $\left.\frac{\partial q_k}{\partial W}\right|_W$, is either undefined or $0$. This issue is well-known in quantization-aware training, which aim to minimize an objective $L(q_k(W))$ with respect to $W$, where $W_q \longmapsto L(W_q)$ is the learning loss. Backpropagating the gradient using the chain rule 
\begin{equation*}
    \label{eq:grad_loss}
    \left.\frac{\partial L\circ q_k}{\partial W}\right|_{W}
    = \left.\frac{\partial L}{\partial W_q}\right|_{q_k(W)} \left.\frac{\partial q_k}{\partial W}\right|_{W}
\end{equation*}
is either not possible or results in a null gradient in this context.

To address this issue, backpropagation through the quantizer is performed using the straight-through estimator (STE) \cite{hinton2012nnml, Bengio2013EstimatingOP,10.5555/2969442.2969588}. The STE approximates the gradient using
\[ \left.\frac{\partial L\circ q_k}{\partial W}\right|_{W}
    \approx \left.\frac{\partial L}{\partial W_q}\right|_{q_k(W)}.
\]
When minimizing models that involve $q_k(W)$, we consistently approximate the gradient using the STE and treat $\Max$ as if it were  independent of $W$.

\section{Activation quantization and complexity}\label{add-fullquant-annexe}

In this section, we provide the details of the strategy used in the experiments to quantize activations, make computations using fixed-point arithmetic, and therefore obtain fully quantized RNNs. 

\myparagraph{Representation:} We use classical notations for fixed-point arithmetics. For integers $k\geq 0$ and $l\geq 1$, the set of $l$ bits fixed-point numbers with $k$ bits for the fractional part is denoted
\[\QQ_{l,k} = \frac{1}{2^{k}}\left\lb-2^{l-1},2^{l-1}-1\right\rb \subset \left[-2^{l-k-1},2^{l-k-1}\right[ \subset \RR.
\]
When $l=k+1$, we simply denote
\[\QQk{k} = \QQ_{k+1,k} = \frac{1}{2^{k}}\left\lb-2^{k},2^{k}-1\right\rb \subset \left[-1,1\right[.
\]
Note that the notation $\QQk{k}$ should not be confused with $\calQ_k$, as defined in \eqref{Qk-eq}.

\myparagraph{Multiplications:} With these notations, the result of the multiplication of two fixed-point numbers $q\in \QQ_{l,k}$ and $q'\in \QQ_{l',k'}$ is such that $q.q'\in \QQ_{l+l'-1,k+k'}$. 

Thus, the result of the multiplication of $q\in \QQk{k}$ by $q'\in \QQk{k'}$ is such that $q.q' \in \QQ_{k+k'+1,k+k'}=\QQk{k+k'}$.

\myparagraph{Additions:} We only add fixed-point numbers with the same fractional size $k$. For instance, for $q$ and $q'\in \QQk{k}$ we have $q +q'\in \QQ_{k+2,k}$.

\myparagraph{Link with weight quantization:} 
In \cref{quant-sec}, we consider a number of bits $k\in\NN$ and, for the quantization of the recurrent weights matrix $W\in\RR^{n_h\times n_h}$,  we consider $\alpha_W= \|W\|_{\max}>0$ and the set of possible values for the entries of $q_k(W)$ is included in $\alpha_W \QQk{k-1}^{n_h \times n_h}$. We write $q_k(W) = \alpha_W \widetilde W$, where $\widetilde W \in \QQk{k-1}^{n_h \times n_h}$.

Similarly, for input weights $U\in\RR^{n_h\times n_i}$, we consider $\alpha_U = \|U\|_{\max}>0$ such that $q_{k}(U) = \alpha_U  \widetilde U$, for $\widetilde U \in \QQk{k-1}^{n_h \times n_i}$.

\myparagraph{Input and Hidden state quantization}
The quantized hidden-state $h_t$, for $t \in \lb 1, T \rb$, is encoded using $k_a$ bits, for $k_a \geq 1$. We also consider a fixed $\alpha_h > 0$ such that the quantized hidden-states are $h_t = \alpha_h \tilde{h}_t$, with $\tilde{h}_t \in \QQk{k_a-1}^{n_h}$. Given a fixed $\alpha_h>0$, to avoid the confusion with $q_{k_a}$ whose scaling parameter is variable, we denote by $q^{\alpha_h}_{k_a}(a)$ the closest element of $a\in\RR$, in $\alpha_h \QQk{k_a-1}$. We extend this definition to vectors.

In practice, $\alpha_h$ needs to be large enough so that $\alpha_h \QQk{k_a-1}$ covers the interval of values of the full-precision hidden-state variable entries. We can, however, increase $\alpha_h$ to some extent without sacrificing performance. We will use this possibility later on.

Similarly, we quantize any input $x_t$, for $t\in\lb1,T\rb$ using $k_i$ bits, for $k_i\geq 1$. For simplicity of notations, we still denote $x_t$ as the quantized inputs. Using a fixed scaling factor $\alpha_i>0$, we write $x_t = \alpha_i \tilde x_t$, where $\tilde x_t \in \QQk{k_i-1}^{n_i}$.

Again, $\alpha_i$ can be chosen quite freely. In practice, in the experiments reported in \Cref{exp-sec}, we will use the following values.
\begin{itemize}
    \item For the Copy-task and PTB, since the entries of $x_t$ are either $0$ or $1$, for all $t$, we use $\alpha_i= 2$. Notice that the quantization does not affect the input as soon as $k_i \geq 2$.
    \item For the two pixel-by-pixel MNIST tasks, since the entries of $x_t$ are normalized 8 bit unsigned integer value in $[ 0,1 ]$, we take $\alpha_i=1$. The quantization does not affect the input as soon as $k_i\geq 9$.
\end{itemize}

\myparagraph{Rescaling $q_{k}(U)$:}
It can be shown by induction that, for any real number $\lambda >0$, and for all inputs $(x_t)_{t=1}^T$, the vanilla RNN of parameters $(W,U,V,b_o)$ using ReLU\footnote{We do not provide the details here but this idea can be adapted to modReLU.} has the same output as the vanilla RNN of parameters $(W,\lambda U, \frac{1}{\lambda} V, b_o)$.

Indeed, considering an input $(x_t)_{t=1}^T$, denoting $(h^\lambda_t)_{t=1}^T$ the hidden-state variables when using the parameters $(W,\lambda U, \frac{1}{\lambda} V, b_o)$, and using \eqref{eq:vrnn}, we have $h^\lambda_1 = \lambda h_1$, from which we obtain $h^\lambda_2 = \sigma(W \lambda h_1 + \lambda U x_2 ) = \lambda h_2$, etc.

In fact, we have for all $t\in\lb1,T\rb$, $h^\lambda_t = \lambda h_t$. Using $\frac{1}{\lambda} V$ leads to the announced statement.

In the sequel, we use this idea and instead of applying the network of quantized weights $(q_{k}(W), q_{k}(U), V,b_o) = (\alpha_W \widetilde W, \alpha_U \widetilde U, V, b_o)$, for $\widetilde W \in \QQk{k-1}^{n_h\times n_h}$ and $\widetilde U \in \QQk{k-1}^{n_h\times n_i}$, we take $\lambda = \frac{1}{\alpha_i\alpha_U} $ and equivalently apply the network of parameters $(\alpha_W \widetilde W, \frac{1}{\alpha_i} \widetilde U, \alpha_i\alpha_UV, b_o)$.

\myparagraph{The fixed-point arithmetic recurrence:}
For simplicity of notation, we drop the exponent $\lambda$ and remind that the quantized hidden-state variable is $h_t =\alpha_h \tilde h_t\in \alpha_h \QQk{k_a-1}^{n_h}$, for a fixed value of $\alpha_h$ that we will choose later on, and a quantized input $x_t=\alpha_i \tilde x_t\in\alpha_i \QQk{k_i-1}^{n_i}$. We define the quantized ReLU function by the composition $q^{\alpha_h}_{k_a} \circ \sigma$. 

The  recurrence \eqref{eq:vrnn} using parameters $(\alpha_W \widetilde W, \frac{1}{\alpha_i} \widetilde U, \alpha_i\alpha_UV, b_o)$ and quantized ReLU becomes
 \begin{align*}
       \alpha_h \tilde h_{t} =    h_{t}  = &~ q^{\alpha_h}_{k_a} \circ \sigma \left( \alpha_W \widetilde W h_{t-1} + \frac{1}{\alpha_i} \widetilde U x_{t}\right) \\
      = &~ q^{\alpha_h}_{k_a} \circ \sigma \left(\alpha_W\alpha_{h} \widetilde W\tilde h_{t-1} +  \widetilde U \tilde x_{t} \right) 
 \end{align*}

The matrix-vector multiplications $\widetilde W\tilde h_{t-1}$ and $\widetilde U \tilde x_t$ can be computed using fixed-point multiplications and additions. We leverage the freedom in choosing $\alpha_h$ to ensure that the multiplication by $\alpha_W \alpha_h$ can be performed with a simple bit-shift. More precisely, to perform the Post-Training Quantization of the activation, given the quantized weights, we first compute $\max_h = \max_{t\in\lb 1,T \rb} \|h_t\|_\infty$, for all the full-precision $h_t$ computed for the inputs in the train and validation datasets. We expect the constraint
\begin{equation}\label{hieriun}
\alpha_h \geq {\max}_h
\end{equation}
 to limit the saturation effects of the activation quantization. We finally take for $\alpha_h$ the smallest number satisfying the constraint \eqref{hieriun} such that $\alpha_W\alpha_h$ is a power of $2$. The values of $\alpha_W\alpha_h$ used in the experiments are given in \cref{tab:alpha_activ}.

Finally, all the entries of  $\sigma \left(\alpha_W\alpha_h \widetilde W\tilde h_{t-1} +  \widetilde U \tilde x_t \right)$ belong to finite set whose size depends on $(k,k_a,k_i)$ and 
$\alpha_W\alpha_h$. We can therefore directly compute $\tilde h_{t}$ using a simple look-up table without any floating point computation.

\begin{table}[h!]
  \caption{Value of $\alpha_W$ and $\alpha_h$  for activation quantization across the datasets considered in the experiments of \Cref{exp-sec}, several bitwidths, and for STE-Björck.}
  \label{tab:alpha_activ}
  \centering
  
  \begin{tabular}{l c c c c c c c c c c c c c c}  
    \toprule
    \multirow{2}{*}{}  & weight &  & \multicolumn{2}{c}{\textbf{Copy-task}} &  \multicolumn{2}{c}{\textbf{sMNIST}} & \multicolumn{2}{c}{\textbf{pMNIST}} &  \multicolumn{2}{c}{\textbf{PTB}} \\
     & bitwidth &  & $\alpha_W$ & $\alpha_W\alpha_h$ &$\alpha_W$ & $\alpha_W\alpha_h$ & $\alpha_W$ & $\alpha_W\alpha_h$ & $\alpha_W$ & $\alpha_W\alpha_h$  \\
    \midrule
    \midrule
 
 \multirow{4}{*}{}     & 4 & & $-$  & $-$ & 0.4338 & 2.0 & 0.3661 & 1.0 & 0.1952  & 1.0 \\
     & 5 &  & 0.2651  & 4.0 & 0.3656 & 2.0& 0.3444 & 1.0 & 0.2350 & 1.0\\
      & 6 &  & 0.2818 & 4.0 & 0.4094 &4.0 &  0.3866  & 1.0 &0.1661  & 1.0 \\
     & 8 &  & 0.2609 & 4.0 & 0.4073 & 2.0& 0.6007  & 2.0 & 0.4827 & 1.0  \\
    \bottomrule
  \end{tabular}
\end{table}


\section{Complements on the Copy-task experiments}
\label{copy-appendix}
\myparagraph{Detailed task description:} This task is the same experiment as in \cite{wisdom2016fullcapacity}, based on the setup defined by  \cite{hochreiter1997long,Arjovsky2015UnitaryER}. The Copy-task is known to be a difficult long-term memory benchmark, that classical LSTMs struggle to solve \cite{Arjovsky2015UnitaryER,bai2018empirical,tallec2018can,helfrich2018orthogonal,kerg2019non,bai2019deep}.

Here, input data examples are in the form of a sequence of $T = T_0 + 20$ time steps, whose first~$10$ elements represent a sequence for the network to memorize and copy. We use a vocabulary $V = \{a_i\}_{i=1}^p$ of $p = 8$ elements, plus a blank symbol $a_0$ and a delimiter symbol $a_{p+1}$. Each symbol $a_i$ is one-hot encoded, resulting in an input time series where $n_i=10$ and an output time series where $n_o=9$ ($a_{p+1}$ is not a target value).

An input sequence has its first $10$ elements sampled independently and uniformly from $V$, followed by $T_0$ occurrences of the element $a_0$. Then, $a_{p+1}$ is placed at position $T_0+11$, followed by another $9$ occurrences of $a_0$. The RNN is tasked with producing a sequence with the same number of time steps, $T_0 + 20$, where the first $T_0+10$ elements are set to $a_0$, and the last ten elements are a copy of the initial $10$ elements of the input sequence.

The naive baseline consists of predicting $T_0+10$ occurrences of $a_{0}$ followed by $10$ elements randomly selected from $V$. Such a strategy results in an expected cross-entropy of $\frac{10 \log 8}{T_0 + 20}$. 

\myparagraph{Hyperparameters:} We use, as in \cite{kiani2022projunn}, 512000 training samples, and 100 test samples.

As described in \cite{kiani2022projunn}, for projUNN-D (i.e. projUNN (FP), 
and STE-projUNN strategies), we use Henaff initialization~\cite{henaff2016recurrent}. For all approaches, we use \textit{modReLU} for the activation function $\sigma$ \cite{helfrich2018orthogonal}.


The initial learning rate is $7e-4$ for projUNN-D (i.e. projUNN (FP), 
and STE-projUNN strategies).
A divider factor of 32 is applied for recurrent weights update, as described in \cite{kiani2022projunn}. For STE-Björck and Björck (FP) the initial learning rate is set to $1e-4$.
For all methods, a learning rate schedule is applied by multiplying the learning rate by $0.9$ at each epoch.
We use the RMSprop optimizer for projUNN-D (i.e. projUNN (FP)
and STE-projUNN strategies), applying the projUNN-D algorithm with the LSI sampler and a rank 1 (as described in \cite{kiani2022projunn}). For STE-Björck, 
we use the classical Adam optimizer. Batch size is set to 128. 

The training spanned 10 epochs.

For LSTM (FP), we report the results given by \cite{wisdom2016fullcapacity}, which indicates that in this setting LSTM (FP) remains stuck at the naive baseline.

\begin{figure}[t]
\begin{tabular}{cc}
    \includegraphics[width=0.47\linewidth]{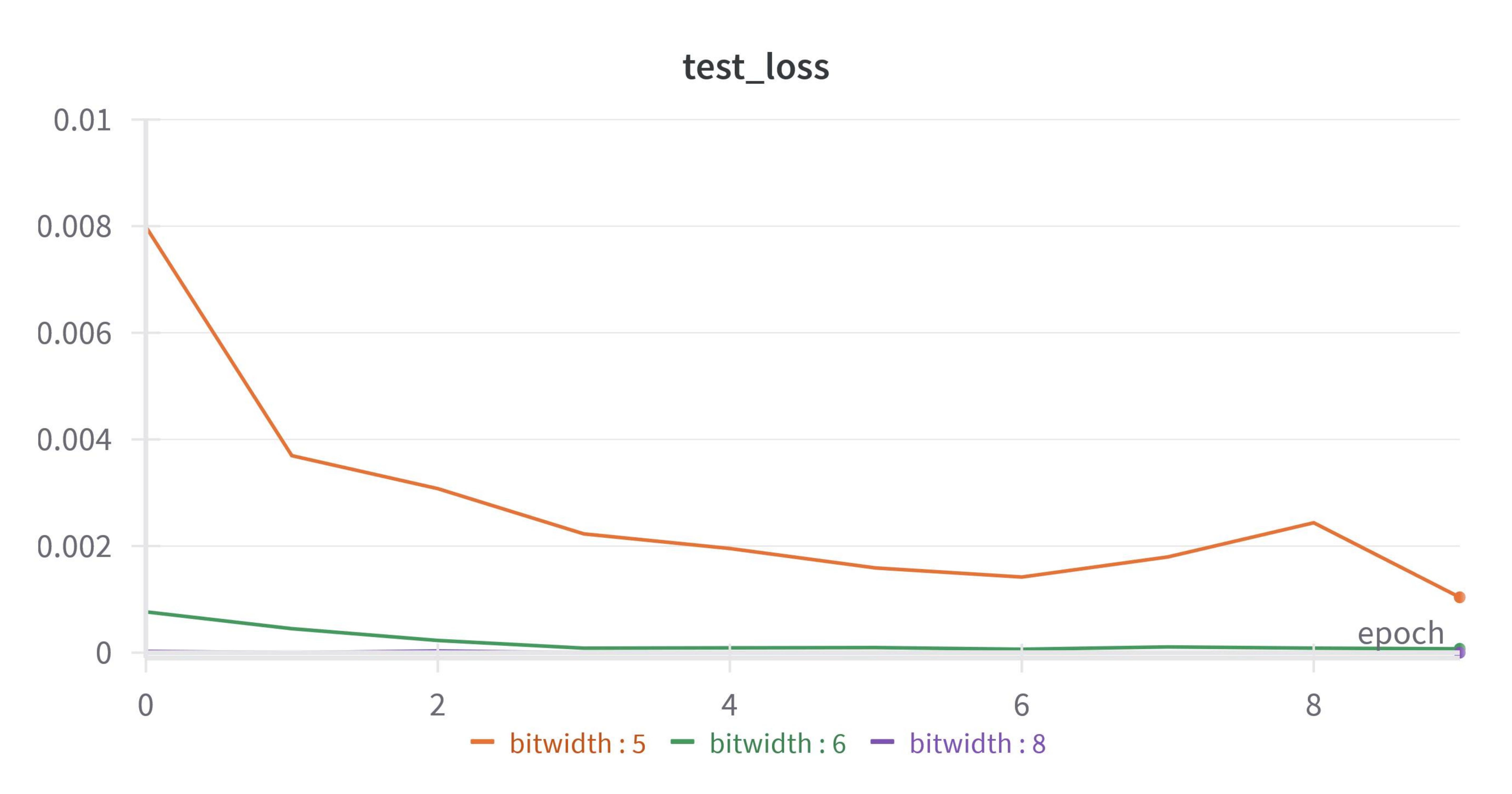}
    &
    \includegraphics[width=0.47\linewidth]{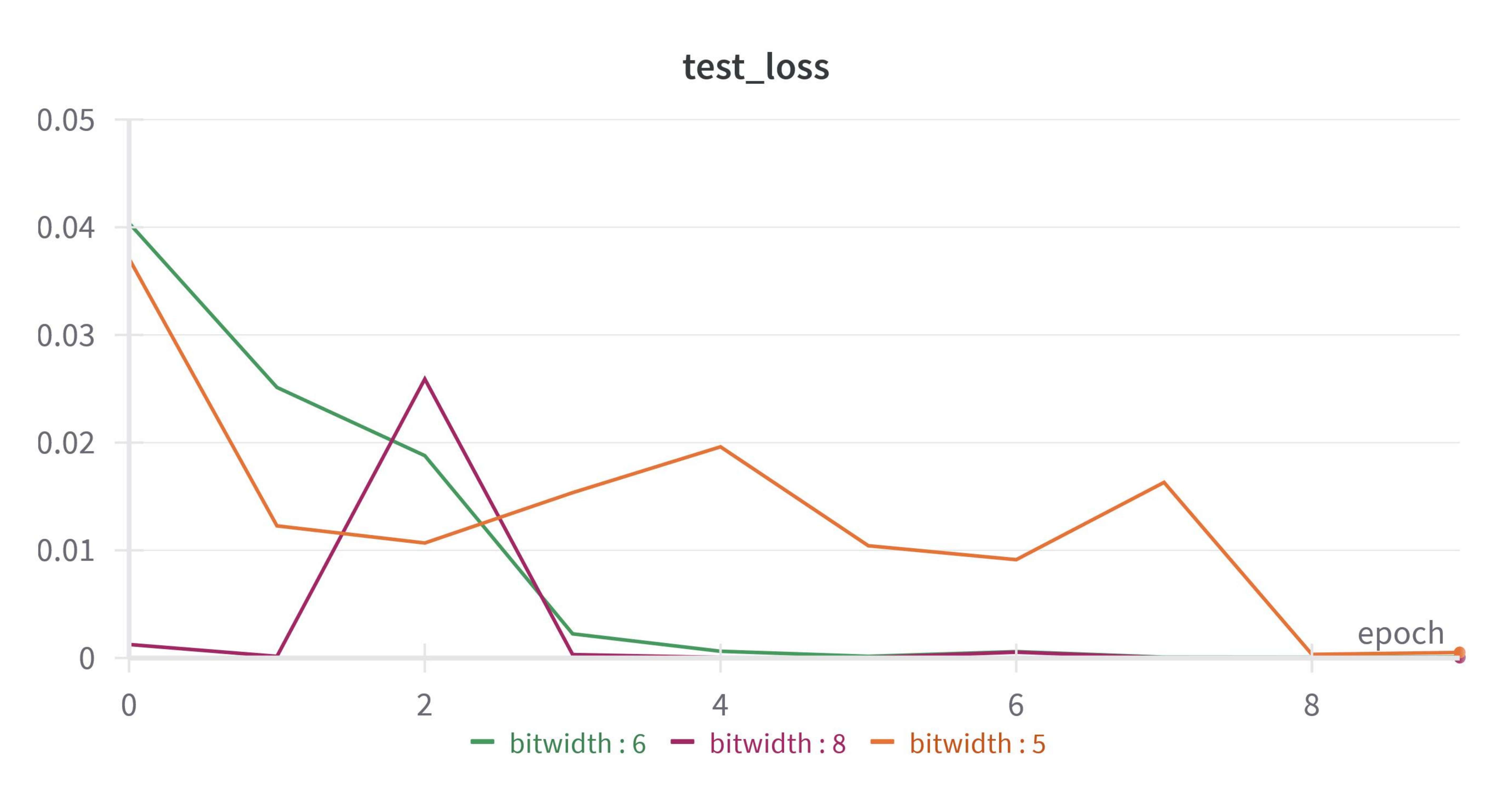}
\end{tabular}
    \caption{Evolution of Test Loss During Training for the Copy-task with $n_h=256$. (Left) STE-projUNN ; (Right) STE-Björck.}
    \label{fig:copy_epochs}
\end{figure}

\myparagraph{Complementary results:} 
We present on \cref{fig:copy_epochs} the evolution of test loss during the training for STE-projUNN and STE-Björck, for several bitwidths.

We also present in \cref{tab:performance_copy_2000} the results for the Copy-task  with $T_0 = 2000$ time-steps and $n_h=256$. This task is more challenging to learn. Both STE-projUNN and STE-Björck need a quantization that uses 6 bits to reach a performance below the naive baseline. It seems likely that increasing the hidden size $n_h$ would allow for a reduction in bitwidth.

\begin{table}[h!]
  \caption{Performance for STE-Björck and STE-ProjUNN  on the Copy-task for $T_0=2000$ and $n_h=256$ for several weight bitwidths.
  }\label{tab:performance_copy_2000}
  \centering
  
  \begin{tabular}{l c c c c c c c }  
    \toprule
    \multirow{2}{*}{Model}    &  \multirow{2}{*}{$T_0$}  & \multirow{2}{*}{$n_h$} & &\multicolumn{4}{c}{weight} \\
    &  &  & &\multicolumn{4}{c}{bitwidth} \\
    \midrule
    &  &  & & FP & 5 & 6 & 8 \\
 \midrule
 \midrule
 \multirow{1}{*}{STE-Björck} & 2000 & 256  &  & 8.9e-6 & NC & 9.4e-3 & 4.3e-05\\
 \midrule
 STE-ProjUNN
     & 2000 & 256 &   & 4.2e-11 & NC & 7.1e-4 & 7.5e-06  \\
    \bottomrule
  \end{tabular}
\end{table}

\section{Complements on the sMNIST/pMNIST task experiments}
\label{mnist-appendix}

\myparagraph{Hyperparameters:} We use the $60,000$ training samples and $10,000$ test samples from the MNIST dataset. 

As described in \cite{kiani2022projunn}, we employ a random orthogonal matrix initialization for the recurrent weight matrix. The activation function $\sigma$ is \textit{ReLU} for STE-Björck, and for PTQ and  STE-pen in the ablation study of \cref{ablation-sec}.
We utilize \textit{modReLU} \cite{helfrich2018orthogonal} for projUNN-D (i.e. projUNN (FP), STE-projUNN strategies), as performances achieved with ReLU are inferior.

The initial learning rate is $1e-3$ for all strategies and weights. It remains constant for STE-pen, in the ablation study. A learning rate schedule is applied by multiplying the learning rate by $0.2$ every $60$ epochs for projUNN-D (i.e. projUNN (FP)
and STE-projUNN strategies) and STE-Björck.

We utilize the RMSprop optimizer for projUNN-D (i.e. projUNN (FP)
and STE-projUNN strategies), implementing the projUNN-D algorithm with the LSI sampler and a rank of $1$ (as described in \cite{kiani2022projunn}). For STE-Björck and STE-pen
we employ the classical Adam optimizer.

For STE-pen, the regularization parameter $\lambda$, which governs the trade-off between optimizing the learning objective and the regularizer (see \cref{eq:ste_pen_obj}), is set to $1e-1$.

The batch size is set to $64$ for STE-pen.
For all other approaches, it is set to 128. 

The training spanned 200 epochs.

LSTM results were taken from \cite{ardakani2018learning} for sMNIST and \cite{kiani2022projunn} for pMNIST.

For FastRNN, results for sMNIST task were given in \cite{kusupati2018fastgrnn}. For pMNIST task, we set the hidden layer size to $n_h=128$, 
using \textit{Tanh} and \textit{Sigmoid} as activation function for recurrent and the gate, as described in \cite{kusupati2018fastgrnn}.

Initial learning rate is set to $1e-3$, and  a learning rate schedule is applied by multiplying the learning rate by $0.7$ every $60$ epochs.
Batch size is also set to 128. We also use the classical Adam optimizer.

\begin{figure}[t]
\begin{tabular}{cc}
       \includegraphics[width=0.47\linewidth]{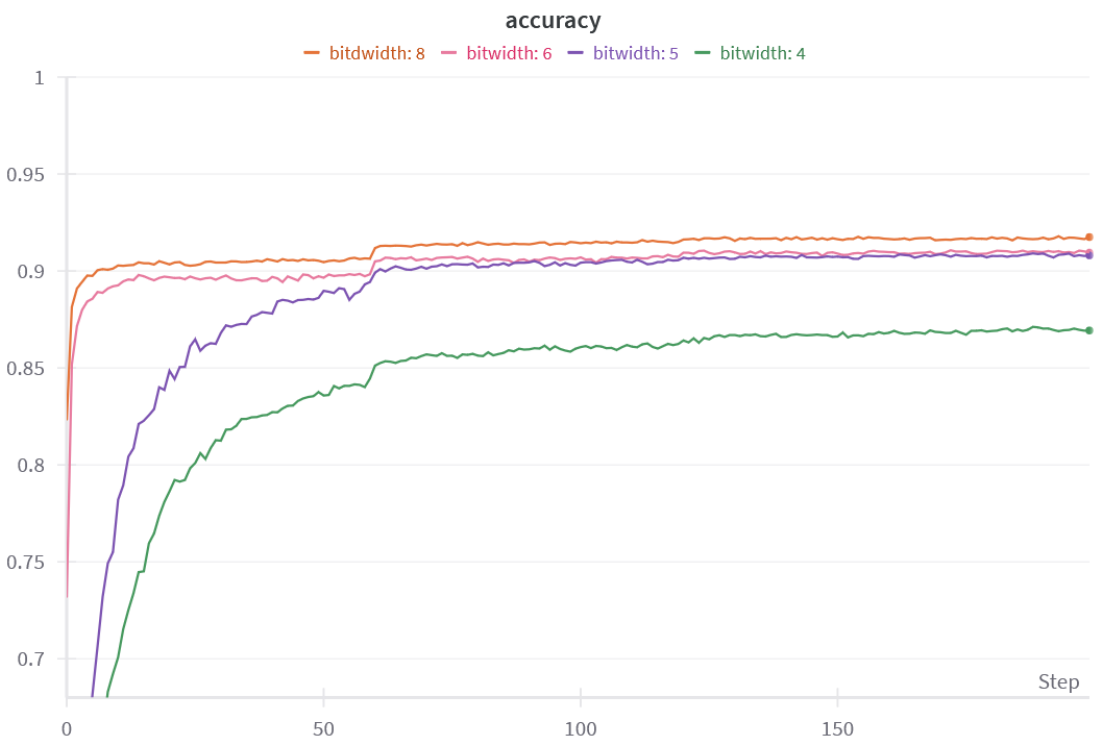}
       &
    \includegraphics[width=0.47\linewidth]{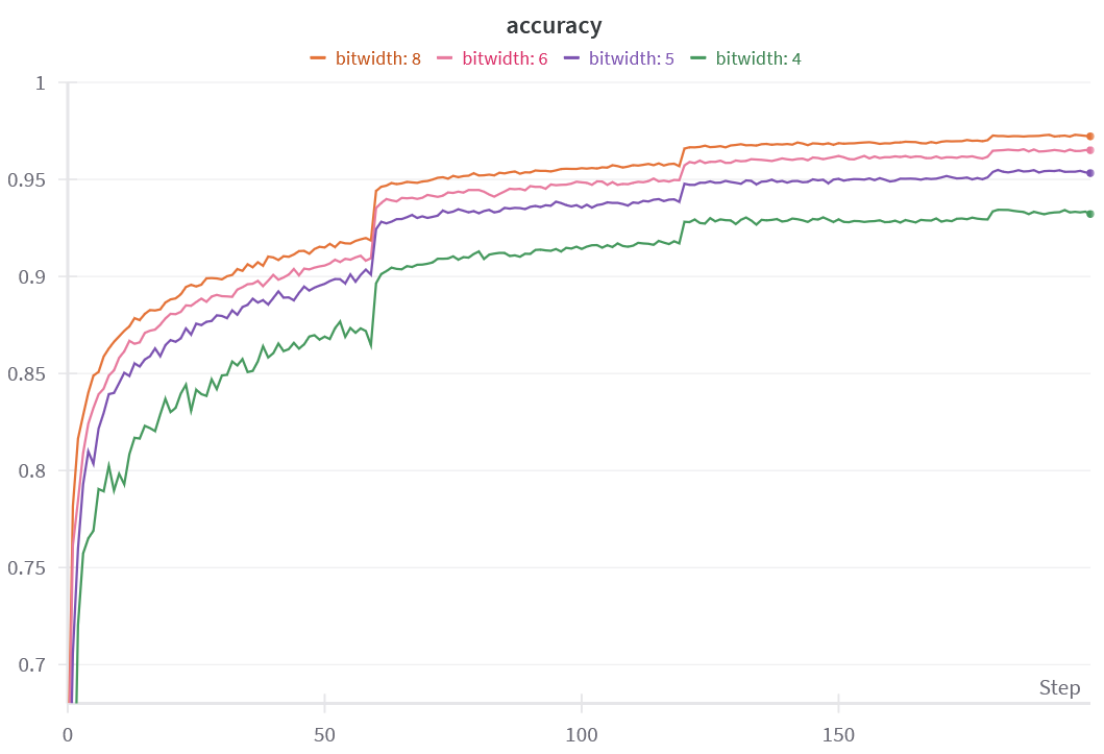}
\end{tabular}
    \caption{Evolution of the accuracy on the training set during training for pMNIST with $n_h=170$. (Left) STE-projUNN ; (Right) STE-Björck.}
    \label{fig:pmnist_loss}
\end{figure}

\myparagraph{Complementary results:} 
We present on \cref{fig:pmnist_loss} the evolution of training
accuracy during the training for STE-projUNN and STE-Björck, for several bitwidths. 

\section{Complements on the  Character level Penn TreeBank experiments}
\label{PTB-appendix}

\myparagraph{Detailed task description:}
The Penn TreeBank dataset consists of sequences of characters, utilizing an alphabet of $50$ different characters. The dataset is divided into $5017$K training characters, $393$K validation characters, and $442$K test characters. Sentences are padded with a blank value when their size is less than the fixed sequence number of time steps  of $150$ characters. The task aims to predict the next character based on the preceding ones. Formally expressed as time series, each sentence represents an input time series of size $n_i=50$ (since characters are one-hot encoded) with $T=150$, and the corresponding output time series is identical to the input time series but shifted by one character.

The models are evaluated using the Bit Per Character measure (BPC), which is the base-$2$ logarithm of the likelihood on masked outputs (to exclude padded values for evaluation).

\myparagraph{Hyperparameters:} As described in \cite{kiani2022projunn}, we take a random orthogonal matrix initialization for the recurrent weights. We use \textit{ReLU} as the activation function $\sigma$ for STE-Björck.
For projUNN-D (i.e., projUNN (FP)
and STE-projUNN strategies), we utilize \textit{modReLU}~\cite{helfrich2018orthogonal} as the activation function, since performances achieved with ReLU were found to be inferior.

The initial learning rate is set to $1e-3$ for all strategies and weight types. 
A divider factor of 8 is applied to recurrent weight updates for projUNN strategies. A learning rate schedule is implemented by multiplying the learning rate by $0.2$ every 20 epochs for all strategies.

We employ the RMSprop optimizer for projUNN-D, utilizing the projUNN-D algorithm with the LSI sampler and a rank 1, as described in \cite{kiani2022projunn}. For STE-Björck, 
we use the classical Adam optimizer.


The batch size is set 
to $128$. 

The training spanned 60 epochs.

For FastRNN, we set the hidden layer size to $n_h=1024$, using \textit{Tanh} and \textit{Sigmoid} as activation function for recurrent and the gate, as described in \cite{kusupati2018fastgrnn}.

Initial learning rate is set to $1e-4$, and  a learning rate schedule is applied by multiplying the learning rate by $0.7$ every $40$ epochs. We use the classical Adam optimizer.
Batch size is also set to 128.  The training spanned 120 epochs.

\section{Complements on the GLUE benchmark}\label{add-glue}

\myparagraph{Detailed task description:}
We consider two binary classification tasks from the GLUE benchmark: SST-2 and QQP. SST-2 is a sentence-level sentiment classification task derived from the Stanford Sentiment Treebank, where each input sentence must be classified as expressing either a positive or a negative sentiment. QQP is a sentence-pair classification task based on Quora Question Pairs, where the objective is to determine whether two questions are semantically equivalent.

For both tasks, input text is tokenized using the \texttt{bert-base-uncased} tokenizer, corresponding to a vocabulary size of $30522$. Each sentence is converted into a sequence of token indices and padded or truncated to a fixed length of $T=128$ tokens. For QQP, the two questions are concatenated into a single sequence, with two separator tokens inserted between them, following the preprocessing used in the reference implementation.

Both tasks are treated as binary classification problems. The network outputs a single scalar logit, i.e., $d_{\mathrm{out}}=1$, and predictions are obtained by applying a sigmoid activation at inference time. The models are trained using the binary cross-entropy loss with logits, and performance is measured using classification accuracy. The final reported scores correspond to the official GLUE evaluation obtained after submitting the predictions of the best validation model to the GLUE benchmark server.

\myparagraph{Hyperparameters:}
For the GLUE experiments, we use the QORNN architecture, composed of a trainable embedding layer followed by a single recurrent layer. Inspired by~\cite{foucault2025hadamrnn}, we slightly change the recurrent layer by applying a linear recurrent unit $h_t = W h_{t-1} + U x_t \in\RR^{n_h}$. The embedding dimension is fixed to $n_i=256$ for both SST-2 and QQP. The recurrent size is fixed to $n_h=64$ for SST-2, and $n_h=512$ for QQP.
Bias terms are used both in the recurrent layer and in the final classification layer. The output is also quantized.

For SST-2, 
the embedding dropout is set to $0.5$, and the model is trained for $60$ epochs with a batch size of $100$. For QQP, the 
embedding dropout is set to $0.1$, and the model is trained for $50$ epochs with a batch size of $20$.

In both cases, the models are optimized with Adam using an initial learning rate of $5 \times 10^{-4}$. A step learning-rate scheduler is applied after every epoch, multiplying the learning rate by a factor of $0.99$. The training uses the binary cross-entropy loss with logits, and the validation metric is binary accuracy.


\section{Complements on the IoT benchmarks}\label{app:iot}

We consider two small-scale, short-term memory IoT datasets: DSA-19 and HAR-2. Both datasets contain sequences of features extracted from sensor-recorded human-motion signals. For HAR-2, the training and test sets contain 7,352 and 2,947 samples, respectively, while for DSA-19, they contain 8,208 and 912 samples, respectively.  

The sequence lengths in DSA-19 and HAR-2 are $T = 125$ and $T = 128$, respectively. In both cases, the associated labels are the corresponding human activities, and the task is to accurately predict the label.

As explained in \cref{har_dsa}, we make HAR-2 a binary classification benchmark, where the networks outputs a single logit, i.e., $d_{\mathrm{out}} = 1$. The models are trained using the binary cross-entropy loss with logits. The DSA-19 benchmark is a $19$-classes classification problem, for which we use the cross-entropy loss. 
In both cases, the performance is measured using the classification accuracy.

\myparagraph{Hyperparameters:} Our experiments are performed using QORNNs with hidden dimension $n_h = 32$. Bias terms are used both in the recurrent layer and in the final classification layer. The activation function $\sigma$ is the Rectified Linear Unit (ReLU).

The initial learning rate is $1e-3$ in both HAR-2 and DSA-19. For DSA-19, we use a learning rate schedule with a decaying factor $0.98$ applied at each epoch. For HAR-2, we use a constant learning rate schedule with decay on validation plateau, with a patience of $10$ epochs. In both cases, we use the classical Adam optimizer.

The number of epochs is $50$ for DSA-19, and 60 for HAR-2.
In both cases, the batch size is set to $50$.


\section{Complements on a regression task: The Adding task}\label{add-sec}

\myparagraph{Detailed task description:}
We consider the Adding task as described in \cite{Arjovsky2015UnitaryER}. In this task, the input to the RNN is a time series $(x_t)_{t=1}^T\in(\RR^2)^T$. Denoting for all $t$, $x_t=(x[0]_t,x[1]_t)$, the sequence $(x[0]_t)_{t=1}^T$ consists of random scalars sampled independently and uniformly from the interval $[0,1]$, while $(x[1]_t)_{t=1}^T$ consists of zeros except for two randomly selected entries set to  $1$. The positions of the first and second occurrences of $1$ are randomly selected, each following a uniform distribution over the intervals $\lb 1,T/2\rb$ and $\lb T/2+1,T\rb$, respectively. The output is the sum of the two scalars from the first sequence, located at the positions corresponding to the $1$s in the second sequence:  $\sum_t x[0]_t \cdot x[1]_t$. As $T$ increases, this task evolves into a problem that requires longer-term memory. Naively predicting $1$ (the average value of the sum of two independent random variables uniformly distributed in $[0,1]$) for any input sequence yields an expected mean squared error (MSE) of $\approx 0.167$, serving as our naive baseline.

\myparagraph{Hyperparameters:} We follow \cite{helfrich2018orthogonal} for most settings and consider $T=750$. We use as in \cite{kiani2022projunn}, 100000 training samples, and 2000 test samples.

As described in \cite{kiani2022projunn}, for all models, 
the activation function $\sigma$ is the Rectified Linear Unit (ReLU), and $\sigma_o$ is the identity. The recurrent weight matrix is initialized to the identity matrix $I$. 

The initial learning rate is $1e-4$ for projUNN-D (i.e. projUNN (FP)
and STE-projUNN strategies), and $1e-3$ for STE-Björck. A divider factor of 32 is applied for recurrent weights update for projUNN, as described in \cite{kiani2022projunn}. A learning rate schedule is applied by multiplying the learning rate by $0.94$ at each epoch.
We use RMSprop optimizer for projUNN-D (i.e. projUNN (FP)
and STE-projUNN strategies) method, applying the projUNN-D algorithm with the LSI sampler and a rank 1 (as described in \cite{kiani2022projunn}). For STE-Björck we use the classical Adam optimizer. 

Batch size is set to 50. 


The training spanned 50 epochs.

Note that, when learning with projUNN, the recurrent weight matrix remains very close to the identity during the learning process. Since the quantization of such matrices would result in reverting to the identity matrix, we have modified the quantization scheme for this experiment with STE-projUNN strategies . The quantized matrix is defined as $W_q = I + q_k(W-I)$.


\myparagraph{Results:} 
In \cref{tab:adding_all}, we see the results for LSTM (FP) are aligned with those reported in \cite{helfrich2018orthogonal}, demonstrating its capability to learn this task even over $750$ time steps.






\begin{table}[h!]
  \caption{Performance for STE-Björck and STE-ProjUNN on  Adding-task $T=750$ with $n_h=170$ and several weight bitwidths (naive baseline is 0.167).
  }\label{tab:adding_all}
  \centering
  
  \begin{tabular}{l c c c c c c c c }  
    \toprule
    \multirow{2}{*}{Model}    & \multirow{2}{*}{$n_h$} & &\multicolumn{5}{c}{weight} \\
    &  &   &\multicolumn{5}{c}{bitwidth} \\
    \midrule
    &  &   & FP & 2 & 3 & 5 & 6 & 8 \\
 \midrule
 \midrule
 \multirow{1}{*}{LSTM}  & 170  &  & 1.0e-4 &   &  &  &  & \\
 \midrule
 \midrule
 \multirow{1}{*}{STE-Björck}  & 170  &  & 9.0e-3 &  0.153 & 0.065 & 0.040 & 0.034 & 8.8e-3\\
 \midrule
 STE-ProjUNN
      & 170 &   &2.0e-4 & 0.170 & 0.165 & 0.080 & 0.147 & 0.062\\
    \bottomrule
  \end{tabular}
\end{table}

In \cref{tab:adding_all}, we observe that both STE-Björck and STE-projUNN achieve a lower MSE
than the naive baseline, even with only 2 or 3 bits.
However, the STE-projUNN strategy is more challenging to learn than STE-Björck. This is possibly due to the resulting matrices being close to the identity.




\begin{table}[h!]
  \caption{Performance for STE-Björck and STE-ProjUNN on  Adding-task $T=750$ with $n_h=400$ and several weight bitwidths (naive baseline is 0.167).
  }\label{tab:adding400}
  \centering
  
  \begin{tabular}{l c c c c c c c c }  
    \toprule
    \multirow{2}{*}{Model}    & \multirow{2}{*}{$n_h$} & &\multicolumn{5}{c}{weight} \\
    &  &   &\multicolumn{5}{c}{bitwidth} \\
    \midrule
    &  &   & FP & 2 & 3 & 5 & 6 & 8 \\
 \midrule
 \midrule
 \multirow{1}{*}{STE-Björck}  & 400  &  & 5.3e-4 &  0.072 & 0.076 & 0.029 & 0.018 & 4.4e-3\\
 \midrule
 STE-ProjUNN
      & 400 &   &2.0e-4 & 0.164 & 0.167 & 0.083 & 0.097 & 0.043\\
    \bottomrule
  \end{tabular}
\end{table}

We present on \cref{tab:adding400} the test accuracy for a larger hidden size $n_h=400$,  STE-projUNN and STE-Björck. Conclusions are similar to the ones established for $n_h=170$. When compared to the results displayed on \cref{tab:adding_all}, the results of STE-projUNN and STE-Björck almost systematically improve. In particular, STE-Björck significantly beats the naive baseline even for $k=2$.

\end{appendices}


\bibliography{biblio}

\end{document}